%% file: main.tex
\documentclass[table]{gtech}
\PassOptionsToPackage{table, usenames, dvipsnames}{xcolor}
\usepackage{xcolor}
\usepackage{amssymb}
\usepackage{multirow}
\usepackage{bigdelim}
\usepackage{longtable}
\usepackage{tabularray}
\usepackage{wrapfig}
\usepackage[most]{tcolorbox}
\usepackage{url}
\usepackage{float}
\usepackage{datatool}
\usepackage{enumitem}
\usepackage{subcaption} 
\usepackage[justification=centering]{caption}

\RequirePackage{tgpagella} 
\RequirePackage{mathpazo}  
\RequirePackage{inconsolata} 
\usepackage{makecell}

\usepackage{booktabs}
\usepackage{array}    
\newcolumntype{C}[1]{>{\centering\arraybackslash}p{#1}}

\usepackage[utf8]{inputenc} 
\usepackage[T1]{fontenc}    
\usepackage{hyperref}       
\usepackage{url}            
\usepackage{booktabs}       
\usepackage{amsfonts}       
\usepackage{nicefrac}       
\usepackage{microtype}      
\usepackage{xcolor}         
\usepackage{xspace}
\usepackage{amsthm}   
\usepackage{amsmath}  
\usepackage{amssymb}  
\usepackage{bm}      

\theoremstyle{definition}

\renewcommand{\title}[1]{\newcommand{\titlelist}{{\huge\fontfamily{optimistic}\selectfont #1}}}

\newcommand{\ignore}[1]{}

\usepackage{arydshln}
\definecolor{CQColor}{rgb}{0.0,0.0,1.0} 

\usepackage{colortbl}
\usepackage{amssymb}
\usepackage{pifont}
\usepackage{booktabs,multirow}
\usepackage{makecell}
\usepackage{tabulary}
\usepackage{fontawesome5}
\usepackage{bbding}
\usepackage{multicol}

\newlength\savewidth

\title{Every Attention Matters: An Eff{}icient Hybrid Architecture for Long-Context Reasoning}

\author{%
  
  \textbf{Ling Team} \\[0.8em] 

  \textbf{Bin Han, Caizhi Tang, Chen Liang, Donghao Zhang, Fan Yuan, Feng Zhu, Jie Gao, Jingyu Hu, Longfei Li, Meng Li, Mingyang Zhang, Peijie Jiang, Peng Jiao, Qian Zhao, Qingyuan Yang, Wenbo Shen, Xinxing Yang, Yalin Zhang, Yankun Ren, Yao Zhao, Yibo Cao, Yixuan Sun, Yue Zhang, Yuchen Fang, Zibin Lin, Zixuan Cheng, Jun Zhou}  \\
}

\abstract{

In this technical report, we present the Ring-linear model series, specifically including Ring-mini-linear-2.0 and Ring-flash-linear-2.0. Ring-mini-linear-2.0 comprises 16B parameters and 957M activations, while Ring-flash-linear-2.0 contains 104B parameters and 6.1B activations. Both models adopt a hybrid architecture that effectively integrates linear attention and softmax attention, significantly reducing I/O and computational overhead in long-context inference scenarios. Compared to a 32 billion parameter dense model, this series reduces inference cost to 1/10, and compared to the original Ring series, the cost is also reduced by over 50\%. Furthermore, through systematic exploration of the ratio between different attention mechanisms in the hybrid architecture, we have identified the currently optimal model structure. Additionally, by leveraging our self-developed high-performance FP8 operator library--linghe, overall training efficiency has been improved by 50\%. Benefiting from the high alignment between the training and inference engine operators, the models can undergo long-term, stable, and highly efficient optimization during the reinforcement learning phase, consistently maintaining SOTA performance across multiple challenging complex reasoning benchmarks.

}
\date{Oct 22, 2025}
\gtechdata[Hugging Face]{https://huggingface.co/inclusionAI/Ring-flash-linear-2.0}
\gtechdata[Hugging Face]{https://huggingface.co/inclusionAI/Ring-mini-linear-2.0}

\begin{document}
\maketitle

\input{sections/1-intro}
\input{sections/2-arch}

\input{sections/3-optimization}
\input{sections/4-pre-training}
\input{sections/5-post-training}

\input{sections/6-evaluation}
\input{sections/7-conclusion}

\bibliographystyle{assets/plainnat}
\bibliography{main}

\end{document}

%% file: sections/1-intro.tex
\begin{figure}[tbhp]
    \centering 
    \includegraphics[width=0.855\textwidth]{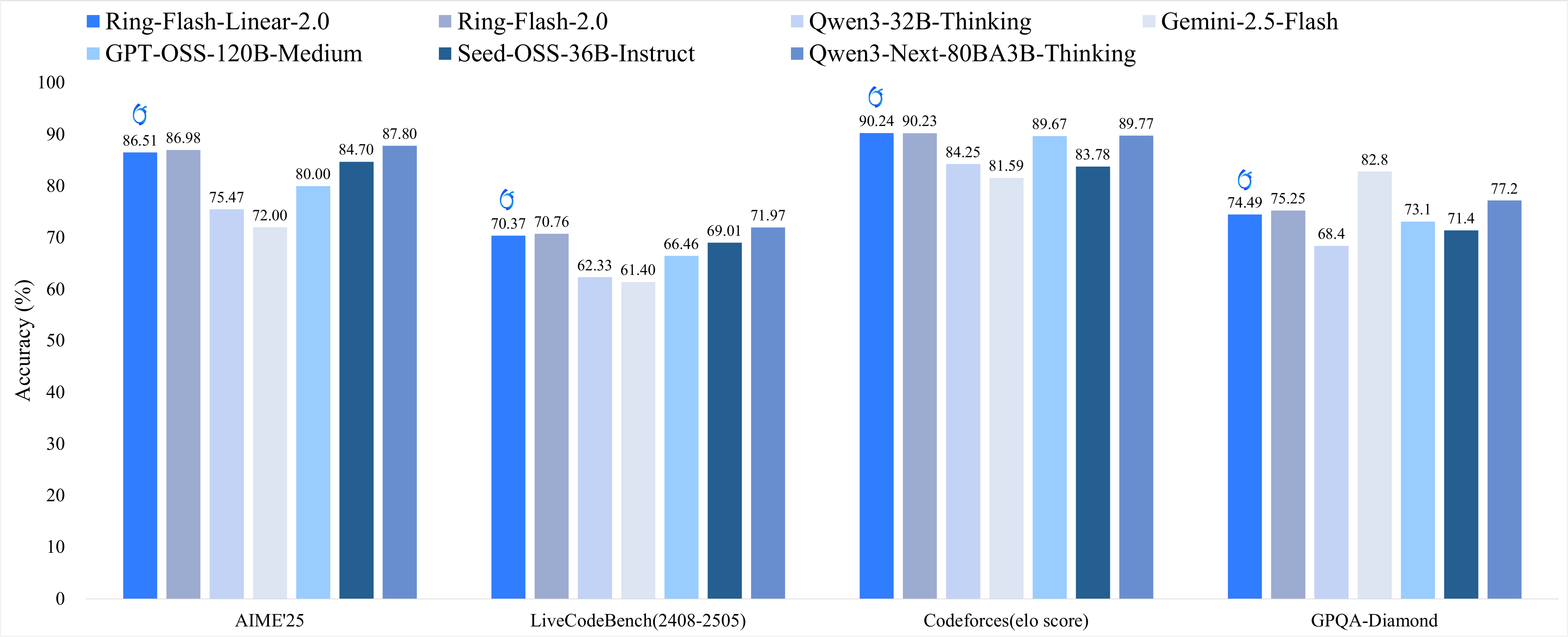} 
    \caption{Benchmark results of Ring-flash-linear-2.0 versus counterparts on representative metrics.} 
    \label{fig:core_bench_flash} 
\end{figure}

\section{Introduction}

In recent years, Test-Time Scaling~\citep{snell2024scaling,muennighoff2025s1} has significantly advanced the capability boundaries of Large Language Models(LLM). Models such as OpenAI’s O-series~\footnote{https://openai.com/index/introducing-o3-and-o4-mini}, DeepSeek R1~\citep{guo2025deepseek}, Gemini 2.5~\citep{comanici2025gemini}, and the Qwen~\citep{yang2025qwen3} Thinking series have achieved enhanced performance by scaling the number of decode tokens. Breakthroughs in reasoning models have also contributed to improvements in non-reasoning tasks. At the same time, growing demands for long-context support in core applications such as Agent systems and code generation have made extended context capability a critical requirement for real-world model deployment.

Amid this progress, two major challenges have emerged: both the training length and output length of language models must be substantially increased, leading to a significant rise in overall resource consumption. Among the contributing factors, the attention mechanism stands out as a key bottleneck. Traditional architectures such as MHA~\citep{vaswani2017attention}, GQA~\citep{ainslie2023gqa}, MQA~\citep{shazeer2017outrageously} and MLA~\citep{liu2024deepseek} exhibit rapidly growing resource overhead as sequence length increases, imposing considerable pressure on both I/O and computation. Specifically, the computational complexity of traditional Softmax Attention grows quadratically with sequence length, while I/O overhead increases linearly with output length. These constraints severely hinder the ability of models to scale to longer contexts.

To address these challenges, research on Linear Attention has progressed rapidly. Approaches such as Retnet ~\citep{sun2023retentive}, Lightning Attention~\citep{qin2023transnormerllm}, Mamba~\citep{gu2024mamba}, Gated Linear Attention~\citep{yang2023gated}, and DeltaNet~\citep{yang2024parallelizing} have been introduced, achieving notable improvements in I/O and computational efficiency: computational complexity is reduced from $O(n^2d)$ to $O(nd^2)$, where $n$ is the sequence length, $d$ is the attention head dimension, and the space complexity of state memory becomes constant. These advances substantially enhance the efficiency of long-text processing and help control inference costs.

However, Linear Attention still exhibits certain limitations in practical applications:

\begin{itemize}
    \item Pure linear language models often underperform in industrial-scale scenarios, particularly as model parameter counts and sequence lengths increase. To mitigate this, hybrid architectures have gained adoption as a balanced approach—retaining a portion of Softmax Attention in the model to maintain both efficiency and expressive power. Models such as Minimax M1~\citep{chen2025minimax}, GPT-OSS~\citep{agarwal2025gpt}, and Qwen3-Next~\footnote{https://huggingface.co/Qwen/Qwen3-Next-80B-A3B-Thinking} have demonstrated promising results using this design.
    \item Although pure Linear Attention offers lower theoretical computational cost, its advantages only become pronounced at sequence lengths beyond 8K. However, the mainstream context length during pre-training typically remains in the 4K–8K range. Moreover, with the growing adoption of Mixture-of-Experts (MoE) architectures—which account for a high proportion of computations at lengths below 8K—the efficiency gains from Linear Attention during pre-training are further constrained.
\end{itemize}

To address these issues, we developed the Ring-linear model series based on a hybrid architecture, include two models: Ring-mini-linear-2.0~\footnote{https://huggingface.co/inclusionAI/Ring-mini-linear-2.0} and Ring-flash-linear-2.0~\footnote{https://huggingface.co/inclusionAI/Ring-flash-linear-2.0}. Our contributions are summarized as follows:
\begin{itemize}
    \item We conducted a systematic study of hybrid linear architectures during pre-training and established a set of effective configuration guidelines.
    \item We implemented infrastructure-level optimizations for both pre-training and inference of hybrid linear models, achieving a 50\% improvement in training performance compare with the native blockwise FP8 mixed-precision training method provided by Megatron~\footnote{https://github.com/NVIDIA/Megatron-LM} and a 90\% increase in inference efficiency.
    \item Through systematic training-inference alignment for reinforcement learning, we significantly enhanced the stability of the model training process, and achieved stable long-horizon RL training.
\end{itemize}

In the remainder of this paper, we first present our model architecture in Section~\ref{sec:intro}. Computational efficiency optimizations during training and inference are detailed in Section~\ref{sec:opt}. We then describe the continued pre-training and post-training procedures in Sections~\ref{sec:pretrain} and~\ref{section_post_training}, respectively. Finally, the performance of our models is evaluated in Section~\ref{sec:evaluation}.

%% file: sections/2-arch.tex
\section{Model Architecture}
\label{sec:intro}

In this section, we first introduce the basic architecture of Ring-linear. Subsequently, we discuss the specific design of the hybrid linear attention mechanism employed, along with several key architecture design choices. To achieve optimal performance within resource constraints and ensure economical training and efficient inference on longer sequences, we conducted extensive scaling law experiments and ablation studies to determine the final design. The key innovation of Ring-linear lies in adopting a highly sparse MoE architecture and utilizing linear attention to the greatest extent possible, as opposed to the traditional softmax attention typically used in standard transformers.


\begin{table}[htbp]
\centering
\resizebox{\textwidth}{!}{%
\begin{tabular}{@{}cccccccccccc@{}}
\toprule
Size  & \begin{tabular}[c]{@{}c@{}}Total \\ Params\end{tabular} & \begin{tabular}[c]{@{}c@{}}Active\\ Params\end{tabular} & \begin{tabular}[c]{@{}c@{}}Non-embedding\\ Active Params\end{tabular} & $n_{layers}$ & $d_{model}$ & $n_{experts}$ & $n_{top\_k}$ & $n_{heads}$ & $n_{kv\_heads}$ & \begin{tabular}[c]{@{}c@{}}Hybrid \\ Ratio\end{tabular} & \begin{tabular}[c]{@{}c@{}}Context\\ Length\end{tabular} \\ \midrule
Mini  & 16.4B                                                   & 1.6B                                                    & 957M                                                                  & 20        & 2048     & 256        & 8         & 16       & 4          & 1:4                                                     & 128K                                                     \\
Flash & 104.2B                                                  & 7.4B                                                    & 6.1B                                                                  & 32        & 4096     & 256        & 8         & 32       & 4          & 1:7                                                     & 128K                                                     \\ \bottomrule
\end{tabular}%
}
\caption{Detailed specs of Ring-linear LLM family of models.}
\label{tab:detailed-specs}
\end{table}

\subsection{Basic Architecture}
\begin{figure}[t]
    \centering
    \includegraphics[
        width=0.78\textwidth,
        trim=0 280 0 0,  
        clip
        ]{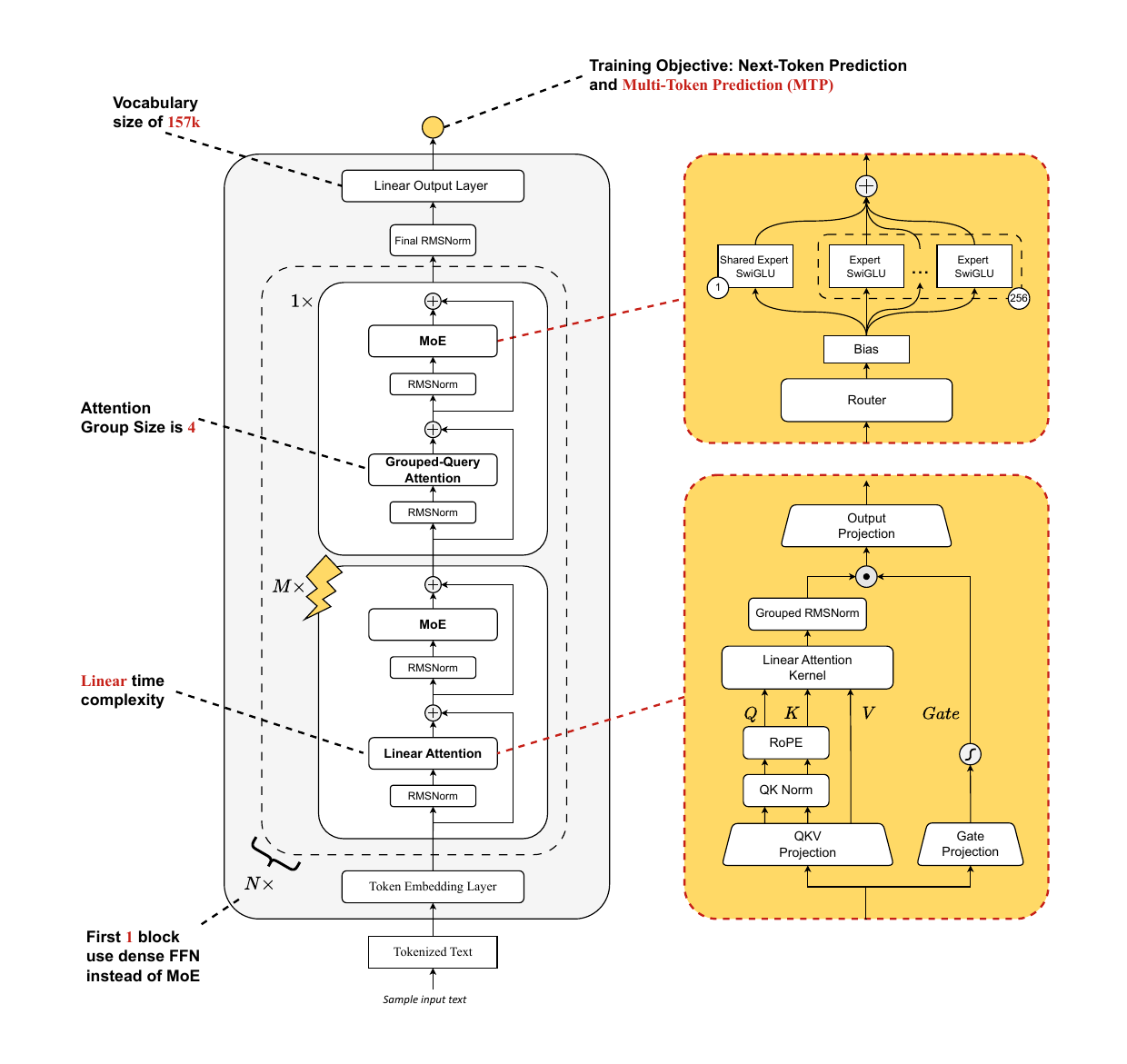}
    \caption{Model architecture of Ring-linear}
    \label{fig:arch}
\end{figure}

As shown in Figure \ref{fig:arch}, Ring-linear is composed of $N$ layer groups, with each layer group consisting of $M$ linear attention blocks and a Grouped Query Attention (GQA) block.
Guided by the principles of Ling Scaling Laws \citep{lingscalinglaws}, Ring-linear incorporates a 1/32 activation-ratio MoE architecture, which has been systematically optimized across a range of design choices. These include expert granularity, the shared-expert ratio, attention balance, an auxiliary-loss-free approach~\citep{wang2024auxiliary} combined with a sigmoid routing strategy, Multi-Token Prediction (MTP) layers, QK-Norm, Partial-RoPE, and other advanced techniques. These innovations enable small-activation MoE models to deliver efficiency improvements of up to seven times compared to their dense counterparts. Specifically, the MLP layer in the first block of the model utilizes a dense MLP instead of a MoE-based MLP.

Currently, two models based on the Ring-linear architecture have been open-sourced. These are the Ring-mini-linear-2.0, with a total of 16 billion parameters and 1.6 billion activated parameters (957M without embedding parameters), and the Ring-flash-linear-2.0, with a total of 104 billion parameters and 7.4 billion activated parameters (6.1B without embedding parameters). The detailed architectural settings of these two models are presented in Table \ref{tab:detailed-specs}.

\subsection{Hybrid Linear Attention}

Although softmax attention has been widely adopted in various LLMs, it suffers from quadratic computational complexity with respect to sequence length and requires Key-Value (KV) cache storage that scales linearly with sequence length. These limitations hinder both training and inference efficiency, especially in long-text scenarios such as test-time scaling and long-context retrieval tasks.
\subsubsection{Linear Attention}
We found that linear attention, which requires constant KV cache storage and incurs computational cost that scales linearly with sequence length, can serve as a viable and significantly more efficient alternative to softmax attention. In Ring-Linear-2.0, we employ a linear attention mechanism with fixed decay ~\citep{sun2023retentive, qin2023transnormerllm} as the foundational implementation for our linear attention. Its mechanism can be formulated as follows:
\begin{equation}
    \textbf{O} = \textbf{Q}(\textbf{K}^{\text{T}}\textbf{V}),
\end{equation}
where \textbf{O}, \textbf{Q}, \textbf{K} and \textbf{V} $\in \mathbb{R}^{n \times d} $ are the output, query, key, and value matrices, respectively, with $n$ denoting sequence length and $d$ representing feature dimension. This linear formulation facilitates recurrent prediction with a commendable complexity of $\mathcal{O}(nd^2)$. 

During the forward pass of Lightning Attention, the output of the $t$-th token can be formulated as
\begin{equation}
    \textbf{o}_{t} = \textbf{q}_{t} \sum_{s \leq t} \lambda^{t - s} \textbf{k}_{s}^{\text{T}} \text{v}_s.
\end{equation}

In a recursive form, the above equation can be rewritten as
\begin{equation}
    \begin{split}
    \textbf{k}\textbf{v}_0 &= 0 \in \mathbb{R}^{d \times d}, \\
    \textbf{k}\textbf{v}_t &= \lambda \textbf{k}\textbf{v}_{t - 1} + \textbf{k}_t^{\text{T}}\textbf{v}_t, \\
    \textbf{o}_t &= \textbf{q}_t(\textbf{k}\textbf{v}_t), \\
    \end{split}
    \label{eq:recurrent}
\end{equation}
where 
\begin{equation}
    \textbf{k}\textbf{v}_t = \sum_{s \leq t} \lambda^{t - s} \textbf{k}_s^{\text{T}} \text{v}_s.
    \label{eq:kv_cache}
\end{equation}
The matrix $\textbf{k}\textbf{v}_t \in \mathbb{R}^{d \times d}$ defined in Equation~\ref{eq:kv_cache} serves as the KV cache in Lightning Attention. In contrast to the linear growth of the KV cache in softmax attention, it requires only constant storage throughout the whole generation process. 

\subsubsection{Hybrid Architecture}


\begin{figure}[t]
    \centering
    \begin{subfigure}{0.585\textwidth}
        \centering
        \includegraphics[width=\linewidth]{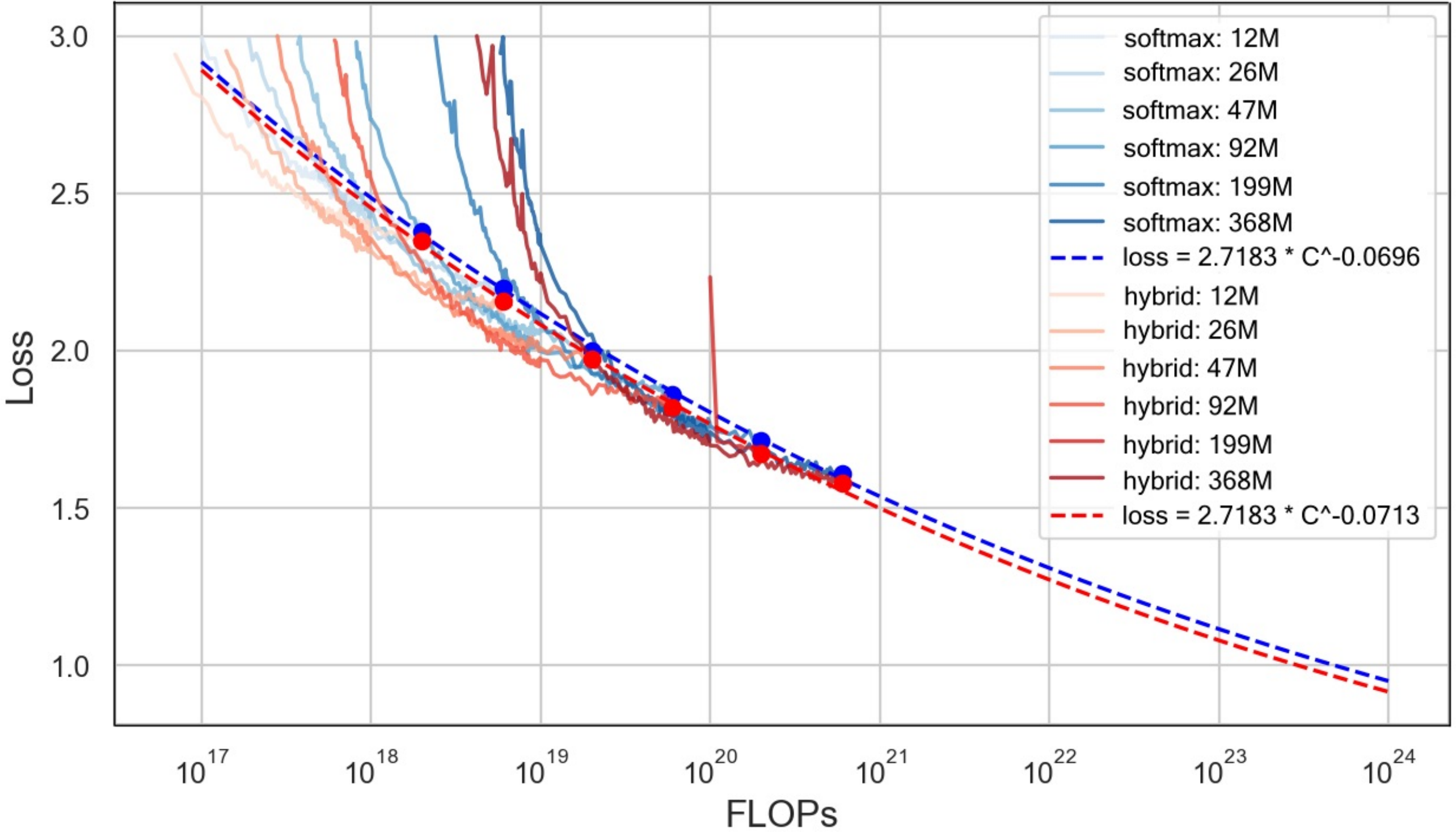}
        \label{fig:sub2}
    \end{subfigure}
    \begin{subfigure}{0.37\textwidth}
        \centering
        \includegraphics[width=\linewidth]{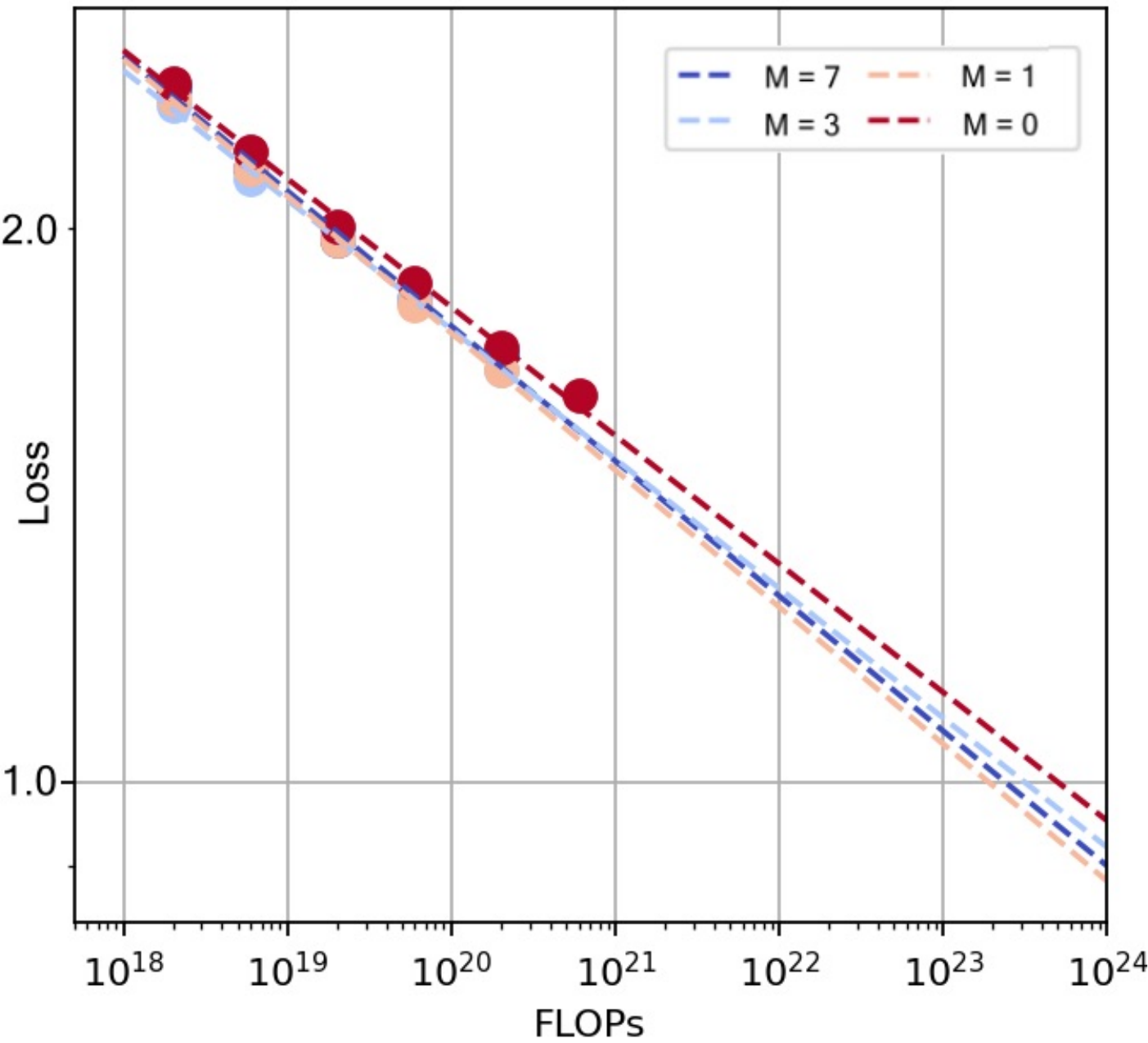}
        \label{fig:sub1}
    \end{subfigure}
    \caption{Scaling law curves. (Left) Curves fitted with different runs of the hybrid linear architecture ($M=1$) and the softmax attention architecture. (Right) Curves comparing different layer group size ($M$) configurations.} 
    \label{fig:scaling_law_all}
\end{figure}

Despite achieving performances comparable to that of softmax attention on most downstream tasks, linear attention underperforms in retrieval. In contrast, the hybrid linear attention architecture not only matches but also surpasses the retrieval and extrapolation capabilities of the pure softmax attention architecture~\citep{li2025minimax}.

We evenly divide the model layers into several groups, each containing layer group size ($M+1$) layers. In each layer group, one softmax attention layer follows $M$ linear attention layers. When $M=0$, the architecture reduces to a pure softmax attention architecture. 

To compare the hybrid linear attention architecture with the softmax attention architecture, we fit scaling law curves for various layer group size configurations. Following Chinchilla's methodology~\citep{hoffmann2022training}, we establish power-law relationships between training FLOP budget and the training loss. The resulting scaling law curves are shown in Figure~\ref{fig:scaling_law_all}. As observed, the hybrid linear architecture consistently outperforms the pure softmax attention architecture. Moreover, a large layer group size (e.g., $M=7$) performs well under high FLOP budgets. Therefore, to strike a balance between efficiency and effectiveness, we set layer group size to 8 ($M=7$) and 5 ($M=4$) for Ring-flash-linear-2.0 and Ring-mini-linear-2.0, respectively.

\subsubsection{Key Architecture Design Choices}
As mentioned earlier, we conducted extensive ablation experiments on the specific implementation details within the linear attention module to determine the optimal architectural design. Our design principle focuses on achieving the best model performance while ensuring efficient distributed training and decoding. Below are some of the key architectural design choices we explored.

\paragraph{Grouped RMSNorm} 
To avoid the all-reduce operations required by a standard RMSNorm layer between the linear attention kernel and output projection under tensor parallelism (size > 1), we employ a grouped normalization strategy. This approach performs RMSNorm locally within each rank, eliminating communication in both the forward and backward passes.

\paragraph{Rotary Position Embedding (RoPE)}
As shown on the right side of Figure \ref{fig:arch}, after performing the normalization operation on the Q and K inputs within the linear attention module, we applied the RoPE operation (which was applied to only half of the dimensions). Our experiments revealed that incorporating RoPE resulted in a reduction of approximately 0.004 in the training language model (LM) loss.

\paragraph{Head-wise Decay}
Our experiments found that the performance of the hybrid linear model is particularly sensitive to the decay coefficient of the hidden state in the Linear Attention mechanism. In the Lightning Attention setting, using a power-law decay rate for the head-wise decay, as opposed to a linear decay rate, resulted in a reduction of approximately 0.04 in the training LM loss. This improvement also had a significant impact on the performance of downstream tasks.

\subsubsection{Cost Analysis for Decoding}
\begin{figure}[t]
    \centering
    \includegraphics[
        width=0.75 \textwidth
        ]{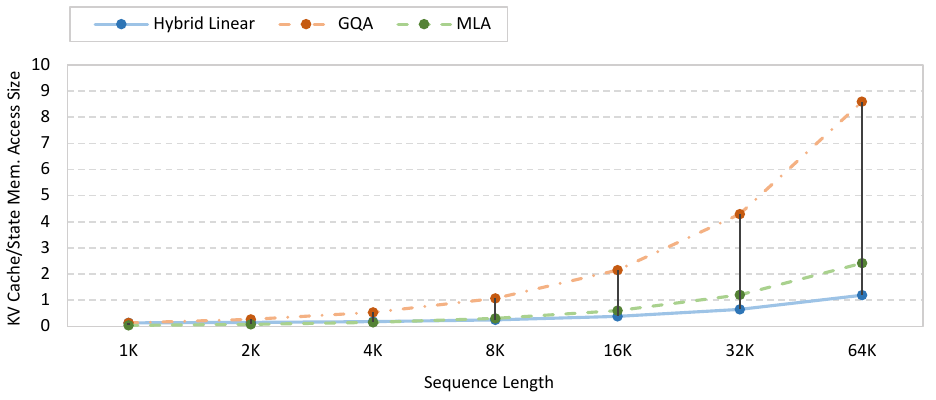}
    \caption{Variation of KV cache/State memory access size with increasing sequence length.}
    \label{fig:decode_cost}
\end{figure}

In practical LLM applications and reinforcement learning (RL) training, decoding efficiency often serves as a critical bottleneck for model test-time scaling. During the LLM decoding process, the limited memory bandwidth of accelerators such as GPUs makes the size of the attention mechanism’s KV cache access a key factor affecting decoding speed. Existing approaches, such as GQA~\citep{ainslie2023gqa} and MLA~\citep{liu2024deepseek}, similarly aim to improve decoding efficiency by optimizing KV cache memory access. In Figure \ref{fig:decode_cost}, we present how the KV cache/State memory access size of our proposed architecture scales with sequence length, along with a comparison against other related methods.

%% file: sections/3-optimization.tex
\section{Computation Optimization}\label{sec:opt}
\begin{figure}[htbp]
    \centering
    \includegraphics[
        width=0.65\textwidth
        ]{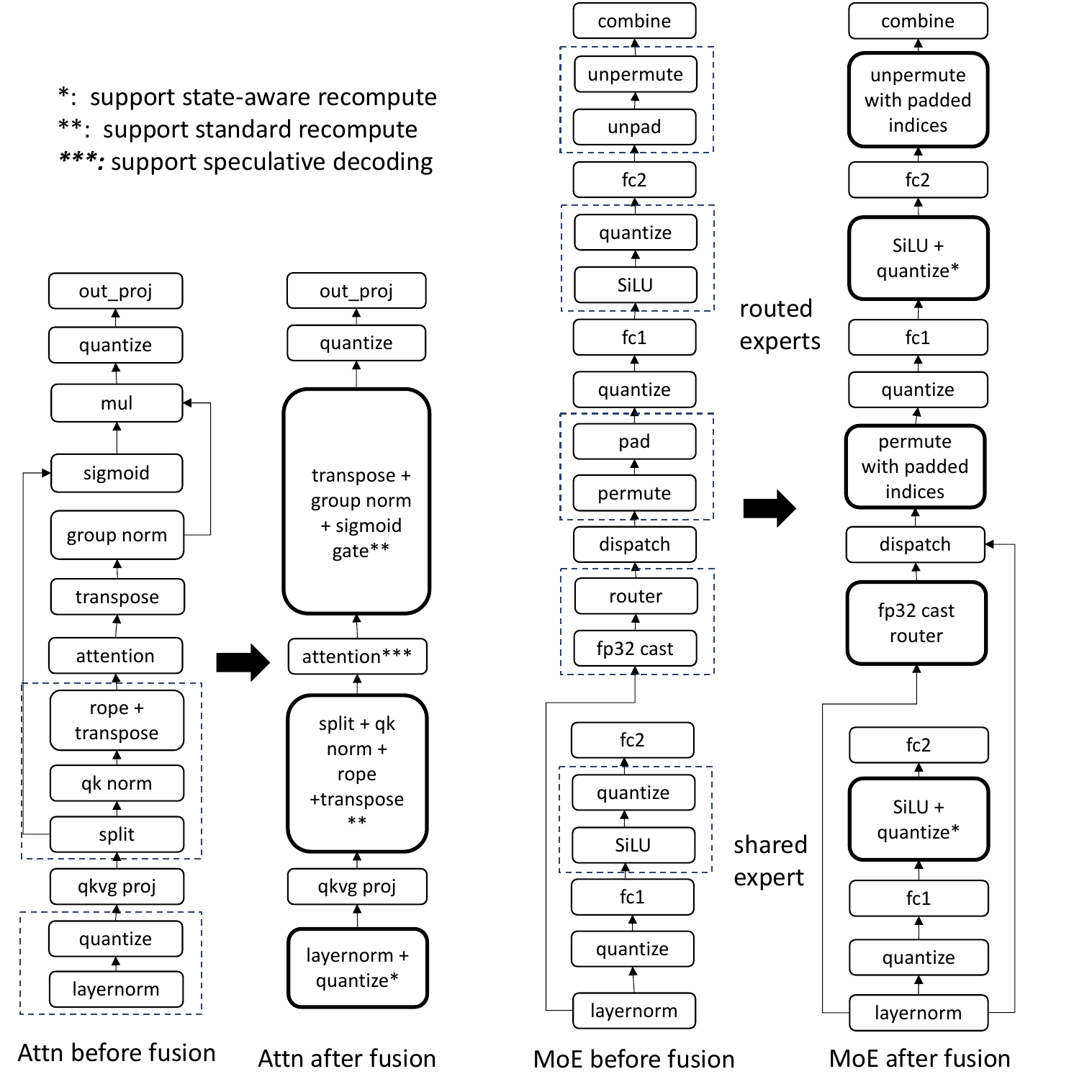}
    \caption{Overview of computation optimization}
    \label{fig:flops_overview}
\end{figure}

To fully demonstrate the computational efficiency advantages of the hybrid linear model, we conducted comprehensive performance optimizations from both training and inference perspectives. The core component of these optimizations lies in kernel fusion and optimization. Additionally, for training, we further implemented kernel fusion and introduced more fine-grained recomputation strategies tailored for the FP8 mixed precision training, significantly improving the training efficiency. On the inference side, we introduced speculative decoding tailored for hybrid linear models, effectively reducing the response latency under low concurrency scenarios.
The overall optimization architecture is illustrated in Figure \ref{fig:flops_overview}.

\subsection{GPU Kernel Optimization}

We performed extensive kernel fusion and optimization for both training and inference. Kernel fusion not only reduces computational latency but also decreases activation memory consumption during training. Lower memory usage enables larger micro-batch sizes. In our experiments, we observed that increasing the micro-batch size from 1 to 2 or from 2 to 4 in MoE training resulted in over 20\% improvement in training efficiency.
We briefly describe several key kernels below; for full implementation details, please refer to our open-source code repository\footnote{linghe: https://github.com/inclusionAI/linghe} \footnote{flood: https://github.com/alipay/PainlessInferenceAcceleration/tree/main/flood}.

\textbf{Linear Gate}: In linear attention layers, we fused the operations related to the gating mechanism (e.g., attention output transpose, group RMS norm, gate sigmoid, and multiplication) into a single kernel. This reduces multiple memory accesses to the GPU memory and lowers activation memory consumption during training.

\textbf{Permute/Unpermute}: Compared to the approach in Megatron, which involves padding/unpadding and permute/unpermute separately, we adopted a more efficient strategy by modifying the routing map directly. This allows us to integrate padding/unpadding into the permute/unpermute operations. Moreover, in extreme cases (e.g., when the number of tokens assigned to a GPU is less than the padding size), this approach does not incur additional overhead.

\textbf{QK Norm + Partial RoPE}: The Ring models incorporate operations such as QK normalization and partial RoPE. We fused the QK normalization with the preceding split, RoPE, and transpose operations.

\textbf{MoE Router}: The baseline approach requires casting hidden states to FP32 before computing the router, which significantly increases both I/O and activation memory. We instead perform cast and compute within the fused kernel, outputting results in FP32 (forward) or BF16 (backward).

\textbf{Linear Attention}: Existing linear attention implementations (e.g., fla.chunk\_simple\_gla\_fwd \footnote{https://github.com/fla-org/flash-linear-attention} ) use 2–4 kernels for the prefill stage, leading to high kernel launch overhead and memory traffic. To address this, we redesigned the partitioning strategy using partitioned Q/K and V, improving performance without sacrificing parallelism. This allows us to use only one Triton kernel (and optionally one additional kernel if Q/K are split).

\textbf{Other Kernel Fusions}: We further optimized kernels based on the model size and parallelization strategy of the Ring model. For example, since we did not use tensor parallelism during training, we were able to develop a more efficient cross-entropy loss kernel without considering distributed loss computation. We also optimized batch computation for operations such as counting zero and calculating gradient norm, achieving significant performance improvements.

\subsection{FP8 training optimization}

Compared to BF16 GEMM, FP8 GEMM offers significantly improved computational speed. However, converting BF16 inputs to FP8 introduces quantization overhead, which becomes a bottleneck, especially on high-performance GPUs like the H800, where quantization can account for nearly 20\% of the GEMM time. Therefore, fusing quantization with adjacent kernels becomes a key strategy to reduce overall latency.

\textbf{Quantization Fusion}: Quantizaiton layers are usually placed after activation or normalization layers. For example, the SiLU is immediately followed by a Linear layer. We modified the SiLU kernel to directly output the quantized tensor, eliminating the need to write and read the BF16 output of SiLU. Assuming the input shape is [M, N], the I/O volume of the non-fused version is approximately $8MN$, while the fused version reduces it to $4MN$. This leads to significant acceleration for I/O-bound kernels. Similarly, in the backward pass, we optimized operations such as SiLU backward to directly output quantized $dy$ and $dy^T$, reducing kernel time by nearly half.

\textbf{State-aware recompute}: We further optimized fine-grained recomputation. In the forward pass, the quantized x is needed to compute $y = x  w$. However, in the backward pass, the transposed quantized $x^T$ is required to compute $dw = x^T  y$. Therefore, the forward pass in recomputation can have different computation and quantization logic compared to the regular forward pass. Leveraging this insight, we designed the forward pass to compute and output only the quantized x when not in recomputation, and only the quantized $x^T$ when in recomputation, thereby reducing redundant computation.

\begin{figure}[t]
    \centering
    \includegraphics[
        width=0.6\textwidth
        ]{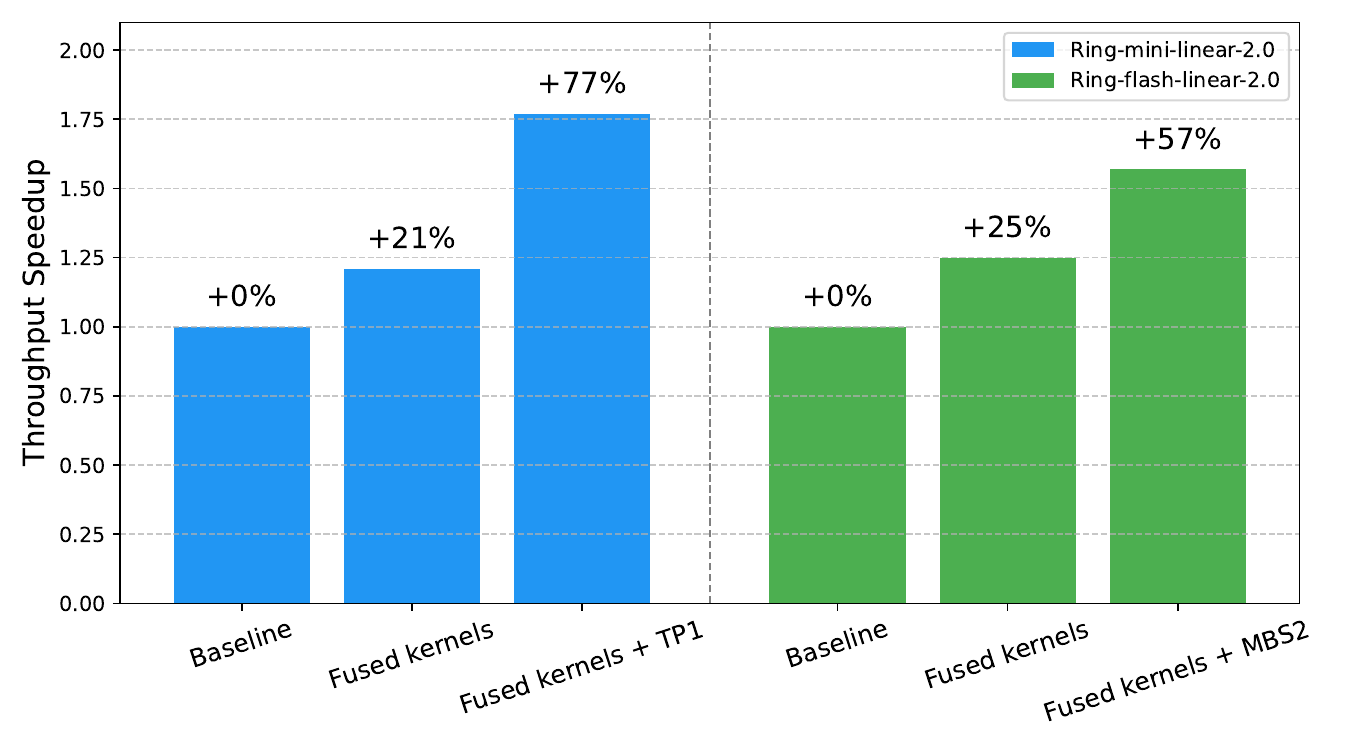}
    \caption{Speedup of LingHe and optimized parameters}
    \label{fig:flops_e2e}
\end{figure}

\subsection{Training efficiency}
The training efficiency with our optimized kernels and parameters was evaluated, as shown in Figure~\ref{fig:flops_e2e}.
For Ring-mini-linear-2.0, the baseline configuration for FP8 blockwise implementation used a global batch size (GBS) of 4416, micro-batch size (MBS) of 4, tensor parallelism (TP) of 2, and 32 H800 GPU. Our optimized version with fused kernels used GBS=4352, MBS=4, and TP=1 (the slightly smaller GBS is due to the inability to reach 4416 with 32 GPUs and MBS=4). Thanks to the memory-efficient fused kernels and more aggressive recomputation, we were able to support MBS=4 with TP=1, resulting in a 77\% improvement in training throughput compared to the baseline.

For Ring-flash-linear-2.0, the baseline configuration used GBS=8352, MBS=1, pipeline parallelism (PP)=6, virtual pipeline parallelism (VPP)=2, and 288 H800 GPUs. Our optimized version with fused kernels achieved a 25\% improvement in training throughput over the baseline. By leveraging the saved memory, we increased the micro-batch size from 1 to 2 and adjusted the pipeline configuration from PP=6/VPP=2 to PP=8/VPP=1, ultimately achieving a 57\% improvement in overall throughput.

\subsection{Inference efficiency}
Although linear attention is theoretically more computationally efficient, its practical implementations have been limited by the absence of support for advanced inference solutions such as SGLang and the vLLM V1 framework. Moreover, the decoding kernel is often fragmented into multiple operations, which further diminishes overall efficiency. To overcome these limitations, we developed a set of optimized fused linear attention kernels and integrated them into both SGLang and vLLM for the Ring-linear models. These improvements not only extend support for a broader variety of inference modes but also substantially enhance system throughput.

\begin{figure}[t]
    \centering
    \begin{subfigure}{0.45\linewidth}
        \includegraphics[width=\linewidth]{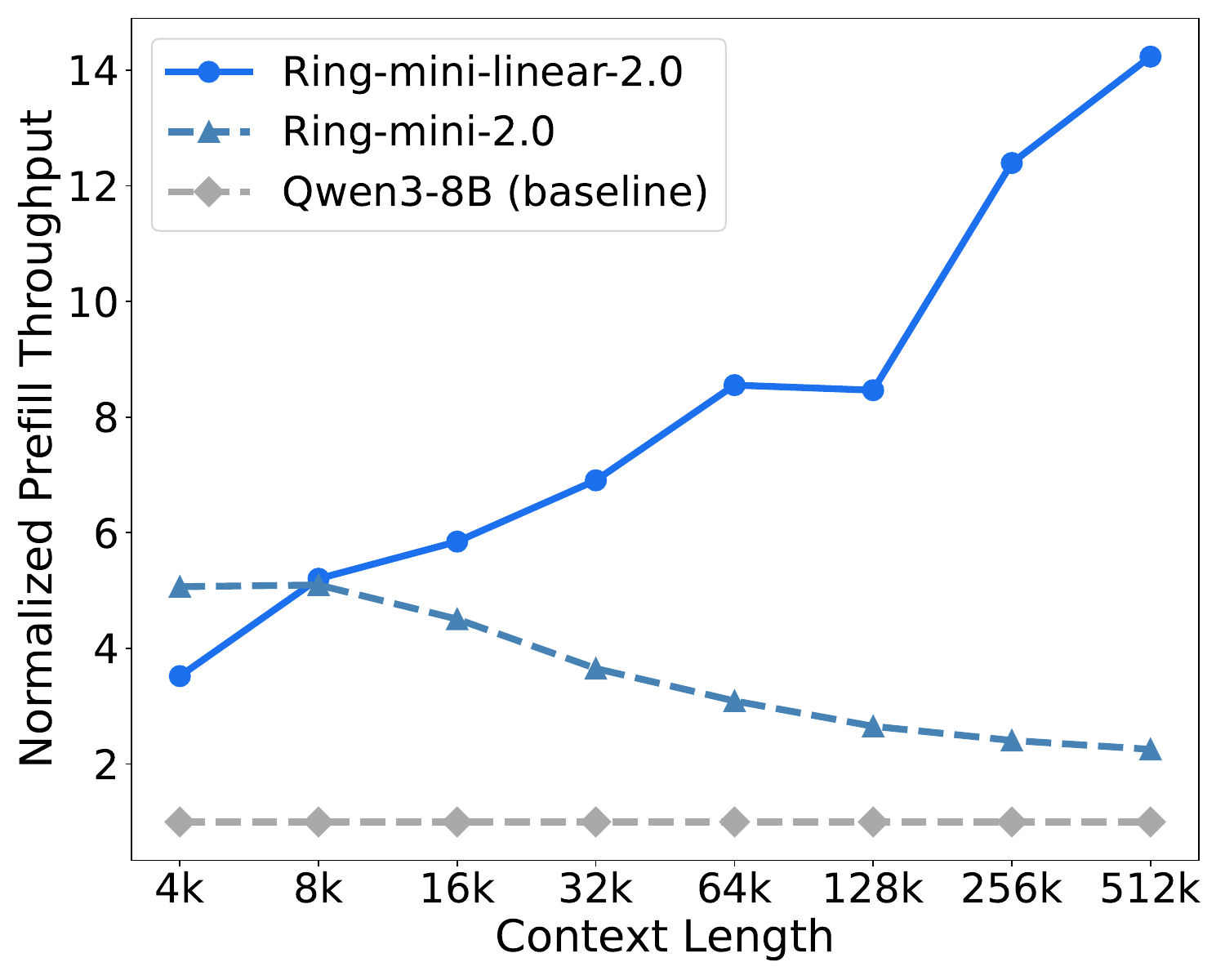}
        \caption{The normalized prefill throughput.}
        \label{fig:throughput_mini_prefill}
    \end{subfigure}
    \hfill
    \begin{subfigure}{0.45\linewidth}
        \includegraphics[width=\linewidth]{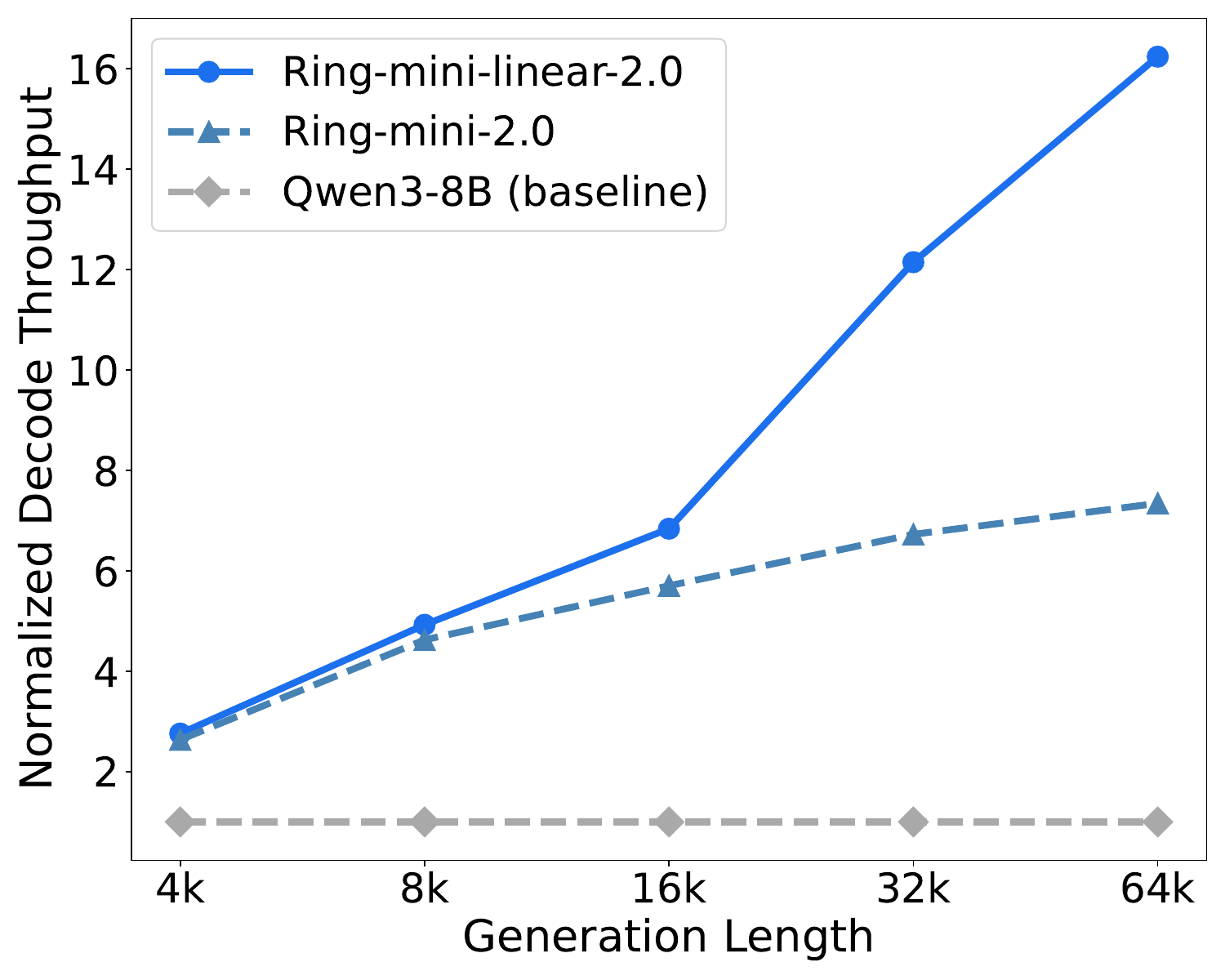}
        \caption{The normalized decode throughput.}
        \label{fig:throughput_mini_decode}
    \end{subfigure}
    \caption{The inference efficiency of Ring-mini-linear-2.0 (TP=1, 1×H20).}
    \label{fig:throughput_mini}
\end{figure}

\begin{figure}[t]
    \centering
    \begin{subfigure}{0.45\linewidth}
        \includegraphics[width=\linewidth]{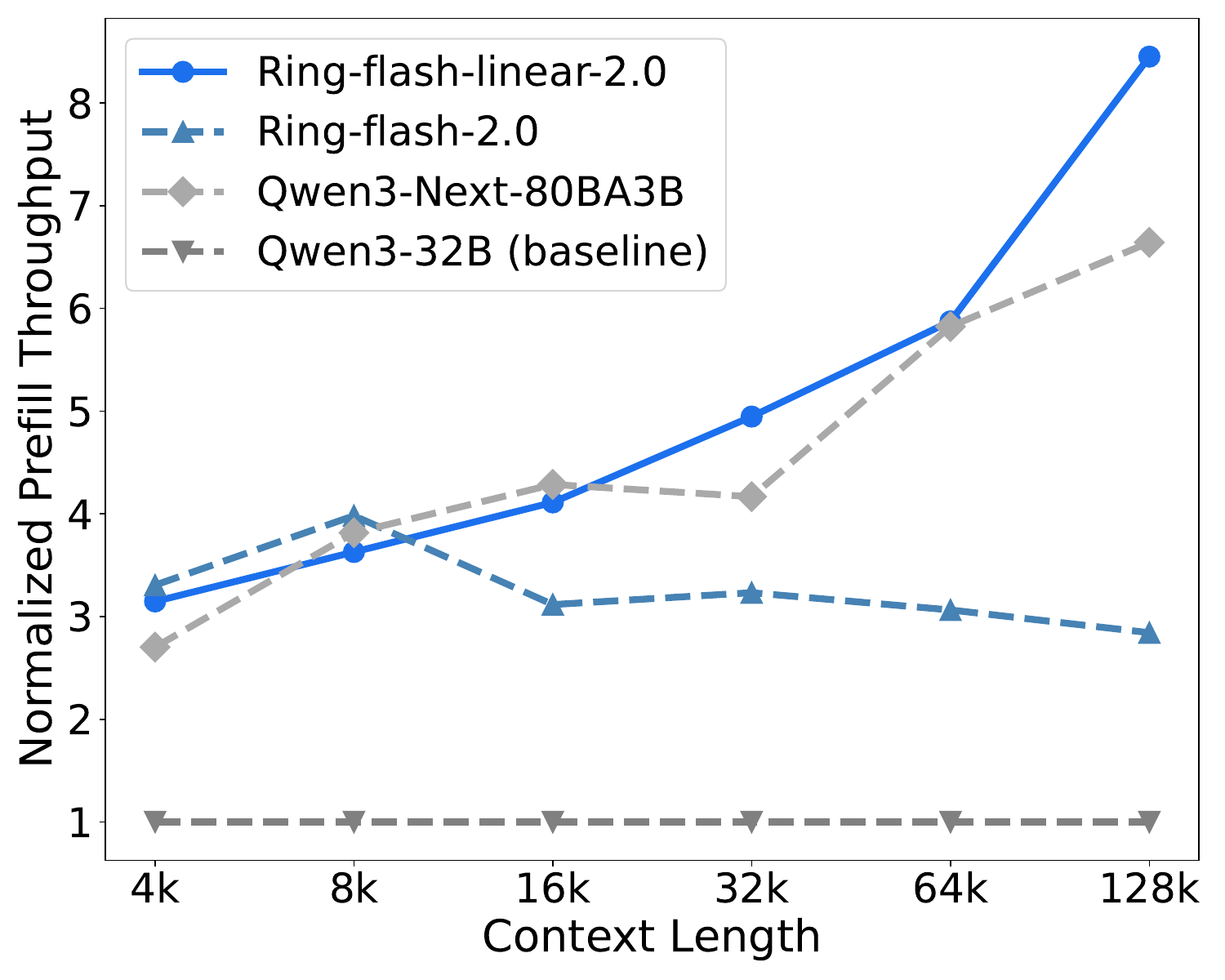}
        \caption{The normalized prefill throughput.}
        \label{fig:throughput_flash_prefill}
    \end{subfigure}
    \hfill
    \begin{subfigure}{0.45\linewidth}
        \includegraphics[width=\linewidth]{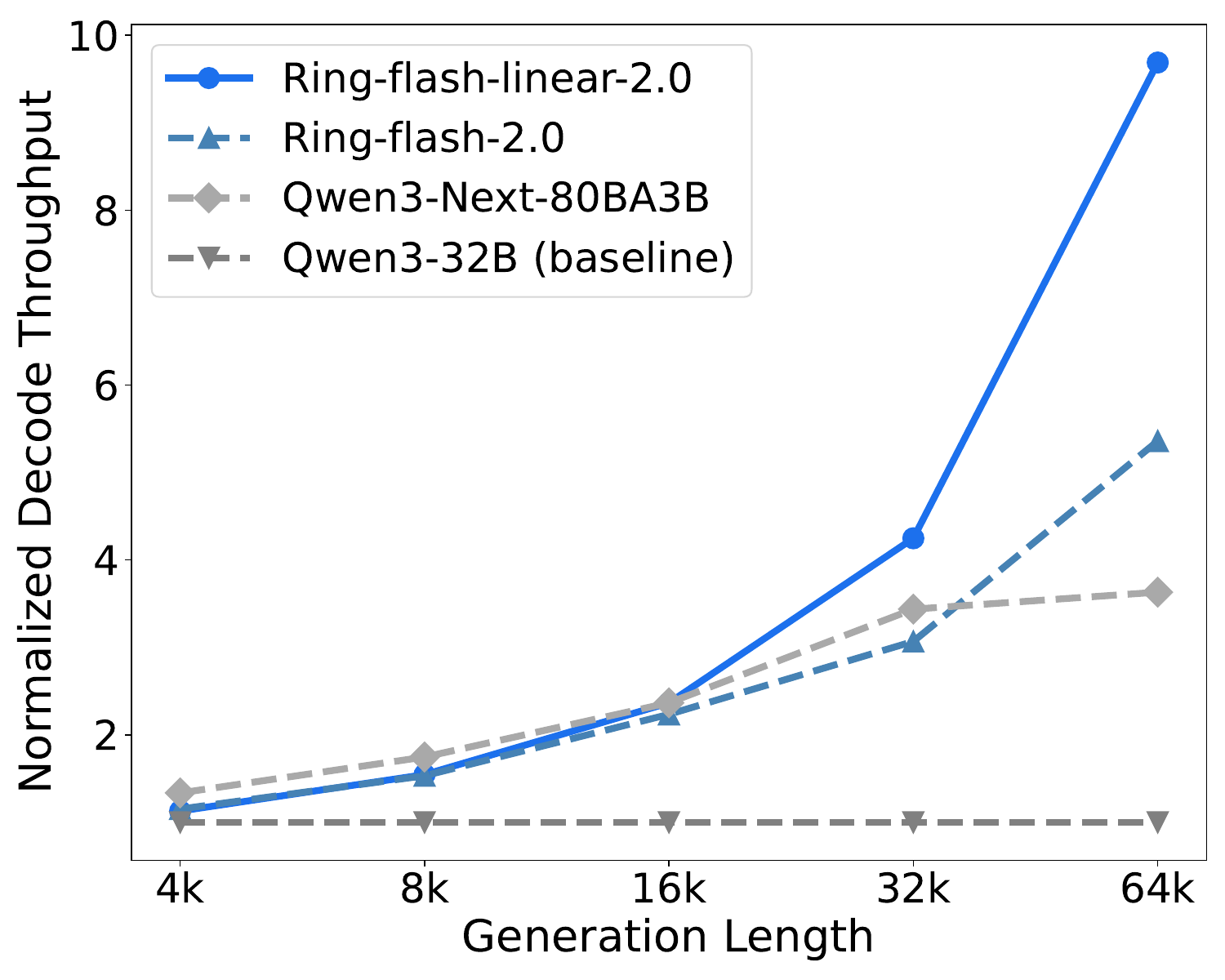}
        \caption{The normalized decode throughput.}
        \label{fig:throughput_flash_decode}
    \end{subfigure}
    \caption{The inference efficiency of Ring-flash-linear-2.0 (TP=4, 4×H20).}
    \label{fig:throughput_flash}
\end{figure}

Using the SGLang framework, we conduct a comprehensive evaluation of the inference efficiency of the Ring-linear models—without MTP—by measuring both prefill and decode throughput. Prefill throughput refers to the number of input tokens processed per second at a batch size of 1, whereas decode throughput denotes the number of output tokens generated per second at a batch size of 64. To facilitate comparison across models, we report normalized throughput relative to a baseline model. The inference performance of Ring-mini-linear-2.0 and Ring-flash-linear-2.0 is presented in Figure~\ref{fig:throughput_mini} and Figure~\ref{fig:throughput_flash}, respectively. To comprehensively illustrate the inference efficiency of Ring-linear models in the figures, we compare them against dense model counterparts (Qwen3-8B and Qwen3-32B~\citep{yang2025qwen3}) and their corresponding Ring-2.0 series (Ring-mini-2.0~\footnote{https://huggingface.co/inclusionAI/Ring-mini-2.0} and Ring-flash-2.0~\footnote{https://huggingface.co/inclusionAI/Ring-flash-2.0}) equipped with softmax attention. Specifically, for Ring-flash-linear-2.0, we conduct an in-depth comparison with Qwen3-Next-80BA3B, which also employs the hybrid linear attention architecture.

During the prefill phase, Ring-linear begins to outperform Ring-2.0 once the context length exceeds 8K, after which its throughput scales rapidly. Ultimately, it achieves more than 2.5 times the throughput of Ring-2.0 and over 8 times that of the baseline models for context lengths beyond 128K. In the decode phase, the Ring-linear models maintain strong performance, surpassing Ring-2.0 when generation length exceeds 4K. At 64K context length, they deliver more than twice the throughput of Ring-2.0 and exceed baseline performance by over tenfold.

Additionally, large language models can benefit from speculative decoding to accelerate inference for small-batch requests. However, existing linear attention kernels in the community do not support custom attention masks, making tree-based speculative decoding infeasible. We developed the first linear attention kernel that supports tree masks and collaborated with our tree-mask-compatible attention kernel to enable speculative decoding for the Ring hybrid linear models. This functionality is already available in our offline inference framework Flood, and we are currently working on porting the kernel to SGLang and integrating it with our previous work lookahead~\citep{zhao2024lookaheadinferenceaccelerationframework}.

%% file: sections/4-pre-training.tex
\section{Continued Pre-Training}\label{sec:pretrain}

To reduce training costs, the base models of the Ring-mini-linear-2.0 and Ring-flash-linear-2.0 models are initialized from Ling-mini-base-2.0-20T~\footnote{https://huggingface.co/inclusionAI/Ling-mini-base-2.0-20T} and Ling-flash-base-2.0-20T~\footnote{https://huggingface.co/inclusionAI/Ling-flash-base-2.0}, respectively. Specifically, the QKV projection in each Linear Attention layer is converted to MHA by expanding parameters along the head dimension, while additional parameters introduced by the gate projection and RMSNorm are randomly initialized. Starting from the above two initialized models, a two-stage continued pre-training strategy is employed to restore the capabilities of the original base models.

\paragraph{Continued training Stage}
This stage samples data from the same corpus used for the Ling-base-2.0-20T models and is trained with 4K context length to restore the base model's capabilities. During this stage, Ring-mini-linear-base-2.0 is trained on 600B tokens, while Ring-flash-linear-base-2.0 is trained on 1T tokens.

\paragraph{Mid-training Stage}
It follows the same strategy as Ling-mini-base-2.0 and Ling-flash-base-2.0. To prepare for the subsequent post-train process, we progressively extend the context window of the models from 4K to 32K and finally to 128K, while correspondingly increasing the proportion of high-quality reasoning data.

\begin{figure}[t]
    \centering
    \begin{subfigure}{0.47\linewidth}
        \includegraphics[width=\linewidth]{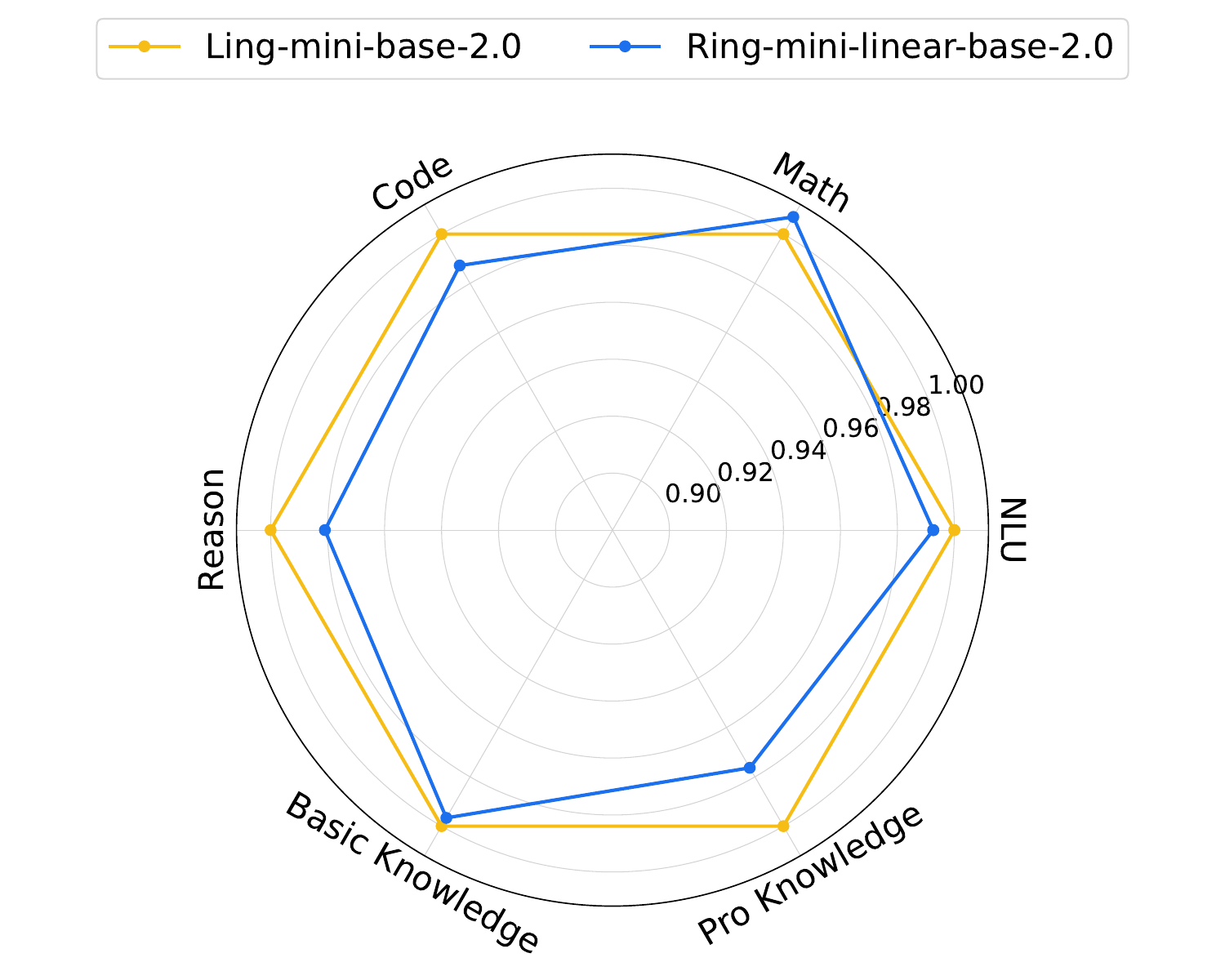}
        \caption{Ring-mini-linear-base-2.0}
        \label{fig:sft_stage1_res}
    \end{subfigure}
    \hfill
    \begin{subfigure}{0.47\linewidth}
        \includegraphics[width=\linewidth]{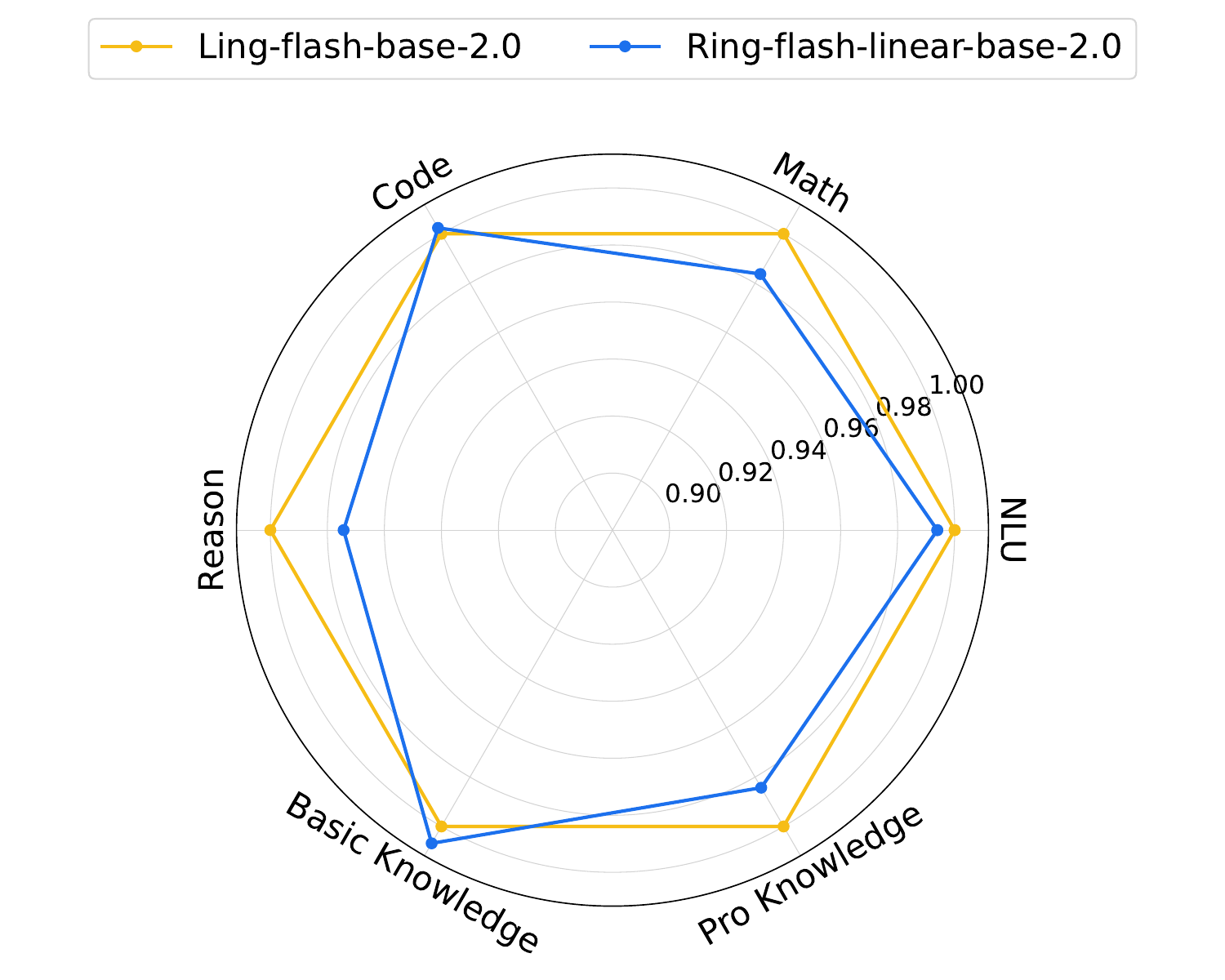}
        \caption{Ring-flash-linear-base-2.0}
        \label{fig:sft_stage2_res}
    \end{subfigure}
    \caption{The normalized performance after continued pre-training.}
    \label{fig:base_res}
\end{figure}

During the above training process, we reuse the critical hyperparameters from Ling-base-2.0 and substitute the standard Warmup-Stable-Decay (WSD) learning rate scheduler~\citep{hu2024minicpmunveilingpotentialsmall} with the Warmup-Stable-Merge (WSM) scheduler~\citep{tian2025wsmdecayfreelearningrate}. By merging mid-training checkpoints to simulate learning rate decay strategy, the resulting Ring-linear-base-2.0 models perform on par with Ling-base-2.0 models trained on 20T tokens across various benchmarks. Figure~\ref{fig:base_res} illustrates the performance of Ring-linear-base-2.0 on different capabilities, including coding, mathematics, reasoning, knowledge, and Natural Language Understanding (NLU). The normalized performance refers to the normalized scores of Ring-linear-base-2.0 relative to Ling-base-2.0. As demonstrated in the figure, Ring-linear-base-2.0 models restore more than 98\% of the original models’ performance in most categories. However, the models exhibit minor deficiencies in reasoning and professional knowledge tasks, which may be attributed to the knowledge forgetting problem~\citep{ibrahim2024simplescalablestrategiescontinually} during the continued pre-training process.

%% file: sections/5-post-training.tex
\section{Post-Training}\label{section_post_training}

Following the continued pre-training process, we further enhance the model's capabilities through a post-training phase, which primarily consists of two stages: Supervised Fine-Tuning (SFT) and RL.
It is noteworthy that we have identified the root cause of the prevalent problem of training collapse in RL: \textit{training-inference disparity}. By achieving systematic alignment between training and inference, we have realized long-term stable RL training.

\subsection{Supervised Fine-Tuning}

The SFT data of Ring-linear places greater emphasis on comprehensive and balanced reasoning capabilities and generalization ability. In addition to including high-difficulty SFT data in areas such as mathematics, coding, science, and logic, the dataset also incorporates general-purpose SFT data covering knowledge, agent tasks, subjective creation, and medical domains. Notably, we have re-synthesized the Function Calling SFT data to better align with more general Function Calling patterns. As a result, Ring-linear series, as reasoning-oriented models, not only perform exceptionally well on reasoning tasks but also provide satisfactory responses to non-reasoning tasks across multiple domains.

All SFT data has undergone stricter de-noising and de-toxification processes, including n-gram filtering and semantic similarity detection, to ensure model safety and the authenticity of its capabilities. 
Following the previous stage, the SFT training process is with context window of 128K.
To prevent overfitting during the SFT phase and to provide ample room for subsequent RL, we selected a checkpoint of an earlier epoch (not the one with the highest benchmark score) for the downstream RL stage.

\subsection{Reinforcement Learning}

Building upon SFT, we subsequently conducted RL across multiple domains including mathematics, coding, science, logic, and subjective tasks. 
All of the samples underwent meticulous screening, from which samples with an appropriate difficulty level were selected for RL training.
In this phase, the model is trained with a sufficiently long context window (e.g., 64K) to strike an optimal balance between performance and efficiency. We observed that with high-difficulty training data, smaller windows (e.g., 32K) introduce potential limitations, namely a high truncation rate and a lower performance ceiling. Conversely, training with a sufficiently long context window proves to be a superior choice, yielding a negligible truncation rate and enabling the model to reach a higher performance ceiling.

Unlike pre-training and SFT, RL for large language models relies on both training and inference. We observe that even standard components in large language models, such as RMSNorm and RoPE, exhibit non-negligible implementation discrepancies across common training (e.g., Megatron, FSDP) and inference (e.g., vLLM, SGLang) frameworks. These discrepancies accumulate and amplify layer by layer, leading to significant differences between training and inference (rollout) outcomes. In extreme cases, the output probability for the same token can be 0 during training and 1 during inference. Such training-inference disparity undermines the theoretical assumption of on-policy, consequently causing instability in RL training and limiting its performance.

Moreover, two prevalent factors notably exacerbate this training-inference disparity:

\begin{itemize}
    \item MoE Architecture: Discrepancies in the selection of experts between training and inference introduce large errors.
    \item Long-CoT Models: Longer model outputs lead to greater cumulative errors.
\end{itemize}

\begin{figure}[htbp]
    \centering 
    \includegraphics[width=0.65\textwidth]{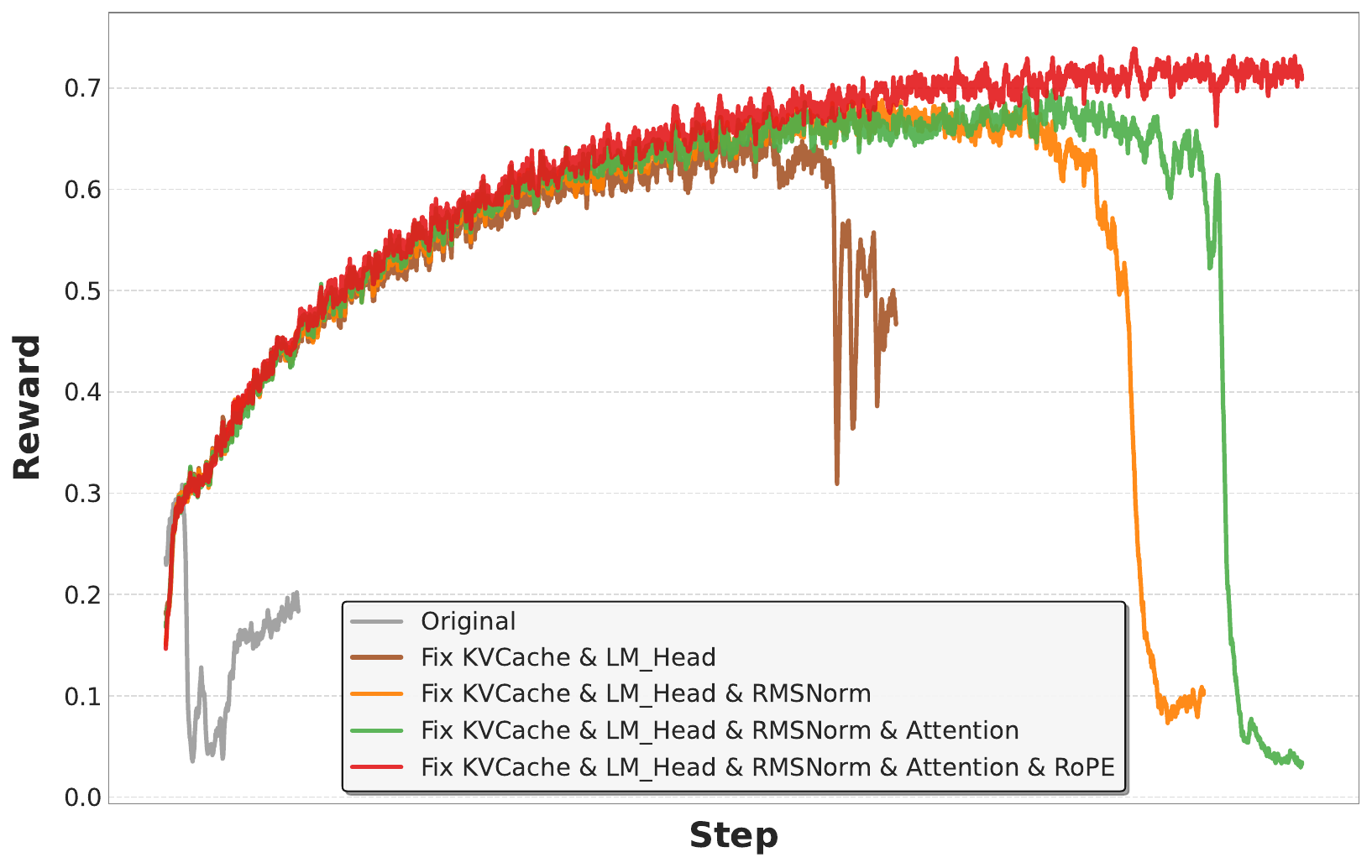} 
    \caption{The ablation experimental results of the RL long-term training curves after correcting the training-inference disparity in different modules of the hybrid linear model.} 
    \label{fig:rl_main} 
\end{figure}

Therefore, the challenges in RL, presented by long-output MoE models and short-output Dense models, are fundamentally different. It is frequently observed that algorithms effective for the latter may not necessarily apply to the former, and vice versa.

Although this issue has recently gained attention in some studies~\citep{zheng2025groupsequencepolicyoptimization,yao2025offpolicy}, most efforts have focused on algorithmic mitigation. In our work, we not only improve RL stability from an algorithmic perspective but also devote substantial effort to systematically addressing the training-inference disparity, aiming to fundamentally solve the problem. As shown in Figure \ref{fig:rl_main}, each aligned module contributes to improved training efficiency and stability in RL.

\subsubsection{Systematic Training-Inference Alignment}
To align training and inference, the same input must be provided to both the training and inference frameworks, with activations compared module-by-module and layer-by-layer. Discrepancies in each activation value should be systematically identified and addressed from front to back. The overall alignment process can be divided into three stages:

\begin{itemize}
    \item Alignment of prefill in training with prefill in inference.
    \item Alignment of prefill in training with decode in inference.
    \item Alignment under different parallelization configurations.
\end{itemize}

To ensure consistency between training and inference, our efforts generally focus on three aspects: ensure identical implementation, maintaining appropriate precision, and eliminating non-determinism. Given the complexity of the full alignment workflow, the following paragraphs highlight specific points within critical modules that are particularly prone to training-inference disparity.

\begin{figure}[htbp]
    \centering
    \begin{subfigure}[b]{0.45\textwidth}
        \centering
        \includegraphics[width=\textwidth]{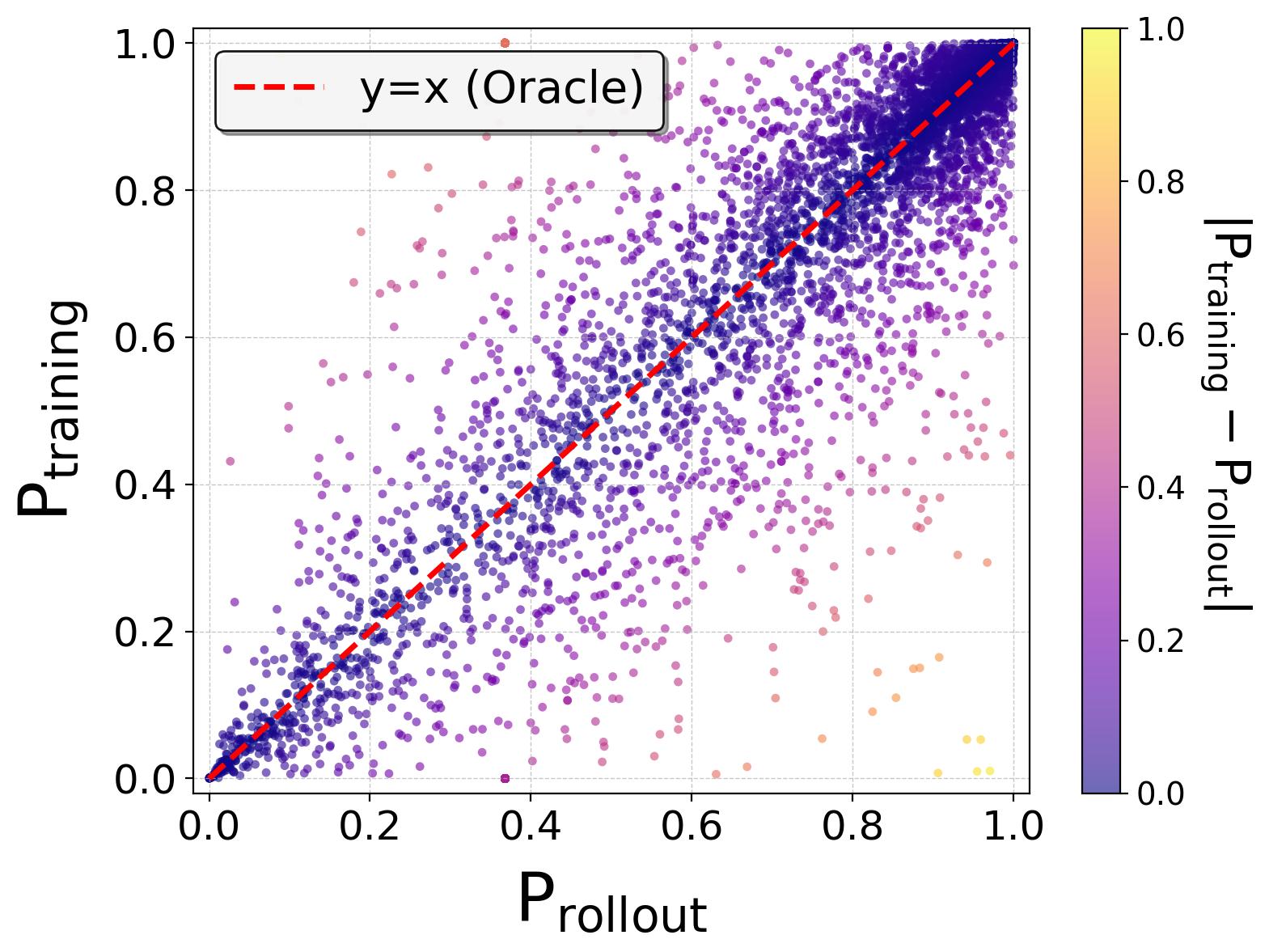}
        \caption{Before}
    \end{subfigure}
    \hfill
    \begin{subfigure}[b]{0.45\textwidth}
        \centering
        \includegraphics[width=\textwidth]{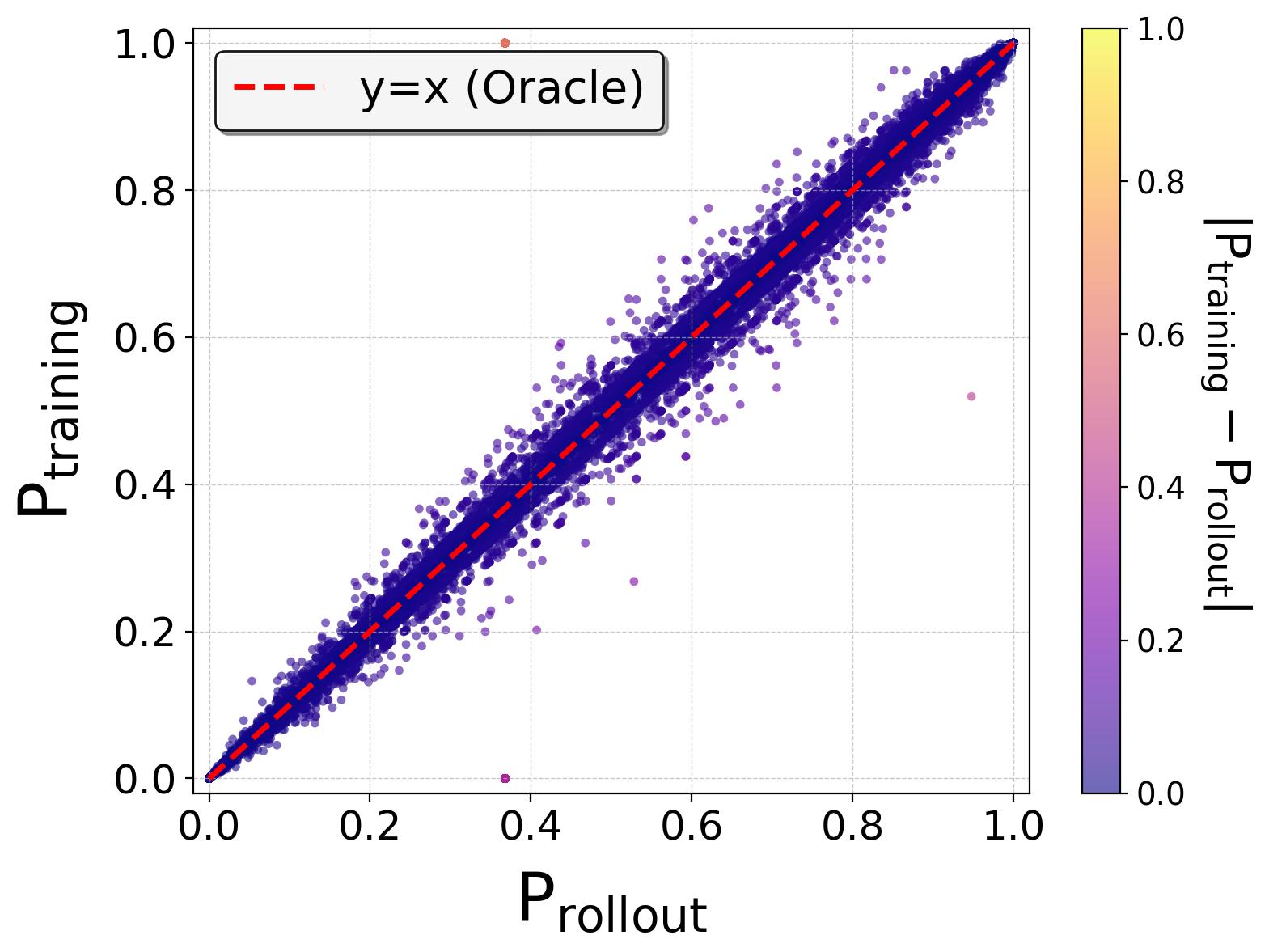}
        \caption{After}
    \end{subfigure}
    \caption{Token output probability distributions before and after KVCache correction.}
    \label{fig:rl_prob_diff}
\end{figure}

\paragraph{KV Cache} In hybrid linear models, the KV states in linear attention require accumulation (Equation~\ref{eq:kv_cache}), which demands higher numerical precision (e.g., FP32). If the KVCache is initialized as BF16 in the inference engine, errors will accumulate progressively during the recurrent process, leading to significant precision divergence, as shown in Figure \ref{fig:rl_prob_diff}.

\paragraph{LM Head} The softmax function is highly sensitive to numerical precision, necessitating FP32 for the \texttt{lm\_head} layer. However, directly configuring the \texttt{lm\_head} to FP32 in the training engine can introduce substantial computational and memory overhead. To address this, we implemented a custom GEMM operator that accepts BF16 inputs and performs conversion and computation within registers. This approach maintains sufficient precision while significantly reducing computational and memory costs.

\paragraph{RMSNorm} The following points require attention: The computation process should use FP32; The epsilon value must be consistent; Keep residual computation in FP32, and un-fuse the RMSNorm and residual in both training and inference.

\paragraph{RoPE} Subtle implementation differences between training and inference should be carefully checked. For example, minor discrepancies often exist between a common PyTorch implementation and a RoPE operator in an inference engine, which can lead to slightly different outputs even with identical inputs.

\paragraph{Attention} First, the used backend must be consistent between training and inference, e.g., FlashAttention~\citep{dao2022flashattention}. Second, watch out for the misalignment between prefill (used during training) and decode (used during inference). This issue causes the training-inference disparity to worsen with longer outputs, making RL training more prone to collapse.

\paragraph{MoE}
In addition to maintaining high precision in router computation, the non-stable \texttt{torch.topk} function must be replaced with a stable implementation. Furthermore, a deterministic order for token permutation and summation is required to prevent discrepancies, and MoE operators must be consistent between training and inference.


\subsubsection{From the Algorithmic Perspective}
When training-inference alignment is achieved at the systematic level, we observe that no additional algorithmic modifications are necessary.
\begin{equation}
    \nabla_\theta\mathcal{J}(\theta) =  \mathbb{E}_{x\sim\textcolor{red}{\pi_{\mathrm{rollout}}}}\Bigl[\nabla_\theta\min\Bigl(  \frac{\textcolor{blue}{\pi_{\mathrm{training}}}(x,\;\theta)}{\textcolor{red}{\pi_{\mathrm{rollout}}}(x,\;\theta_{\mathrm{old}})}\,\hat{A},  \mathrm{clip}\Bigl(    \frac{\textcolor{blue}{\pi_{\mathrm{training}}}(x,\;\theta)}{\textcolor{red}{\pi_{\mathrm{rollout}}}(x,\;\theta_{\mathrm{old}})},    1-\epsilon,\;1+\epsilon  \Bigr)\,\hat{A}\Bigr)\Bigr].
    \label{eq:ppo1}
\end{equation}

Equation \ref{eq:ppo1} described above represents the ideal PPO optimization method, where the true rollout sampling distribution is used for importance sampling weighting. However, due to training-inference disparity, rollout probabilities often differ significantly from training probabilities, making them unsuitable for importance sampling. Most RL frameworks, e.g., verl~\citep{sheng2024hybridflow} and OpenRLHF~\citep{hu2024openrlhf}, resort to recomputing the training probabilities by re-forwarding the rollout data through the training engine, which leads to the following Equation \ref{eq:ppo2}.
\begin{equation}
    \nabla_\theta\mathcal{J}^{\prime}(\theta) =  \mathbb{E}_{x\sim\textcolor{red}{\pi_{\mathrm{rollout}}}}\Bigl[\nabla_\theta\min\Bigl(  \frac{\textcolor{blue}{\pi_{\mathrm{training}}}(x,\;\theta)}{\textcolor{blue}{\pi_{\mathrm{training}}}(x,\;\theta_{\mathrm{old}})}\,\hat{A},  \mathrm{clip}\Bigl(    \frac{\textcolor{blue}{\pi_{\mathrm{training}}}(x,\;\theta)}{\textcolor{blue}{\pi_{\mathrm{training}}}(x,\;\theta_{\mathrm{old}})},    1-\epsilon,\;1+\epsilon  \Bigr)\,\hat{A}\Bigr)\Bigr].
    \label{eq:ppo2}
\end{equation}

However, Equation \ref{eq:ppo2} is biased because it totally ignores the training-inference disparity. Consequently, the most reliable path forward is to use the rollout probabilities directly, as in Equation \ref{eq:ppo1}, but only after the training-inference disparity has been systematically corrected.

\begin{figure}[t]
    \centering
    \begin{subfigure}[b]{0.45\textwidth}
        \centering
        \includegraphics[width=\textwidth]{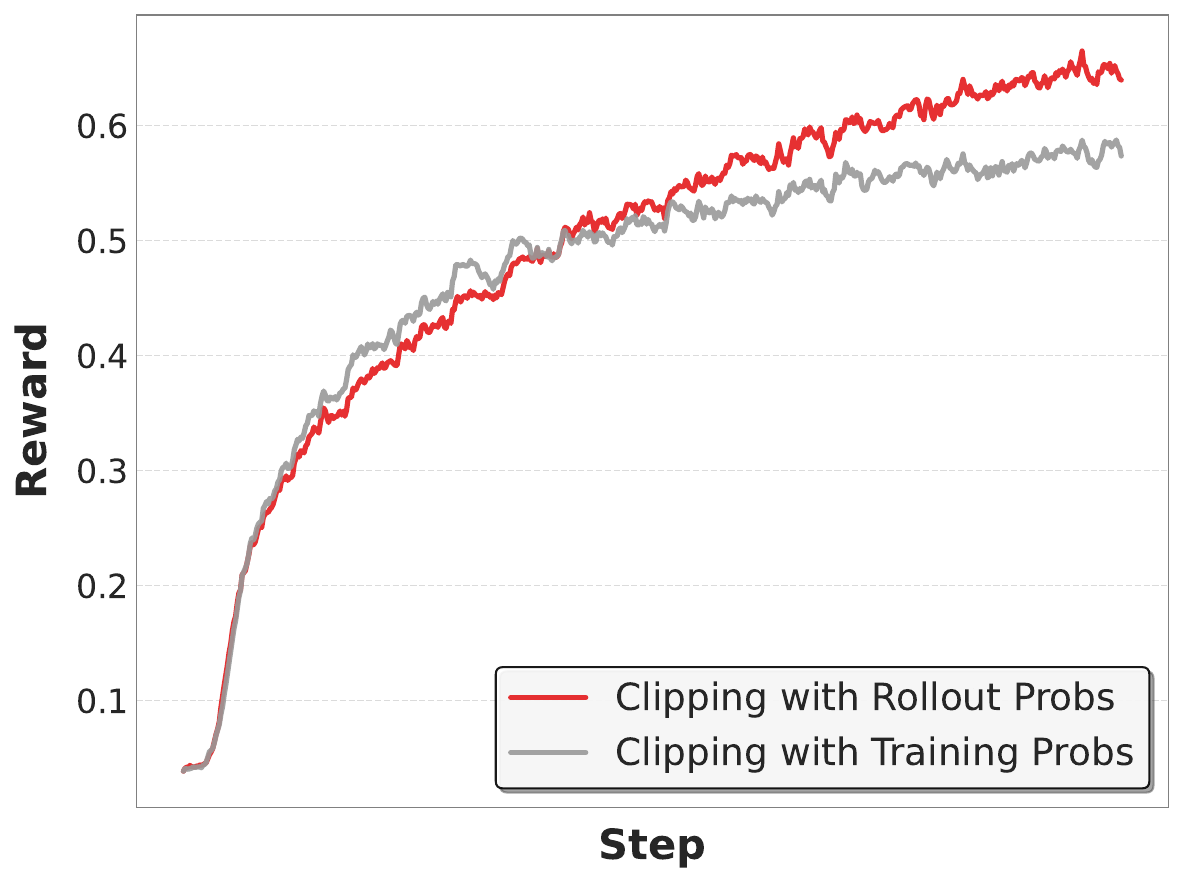}
        \label{fig:rl_alg_compare_reward}
    \end{subfigure}
    \hfill
    \begin{subfigure}[b]{0.45\textwidth}
        \centering
        \includegraphics[width=\textwidth]{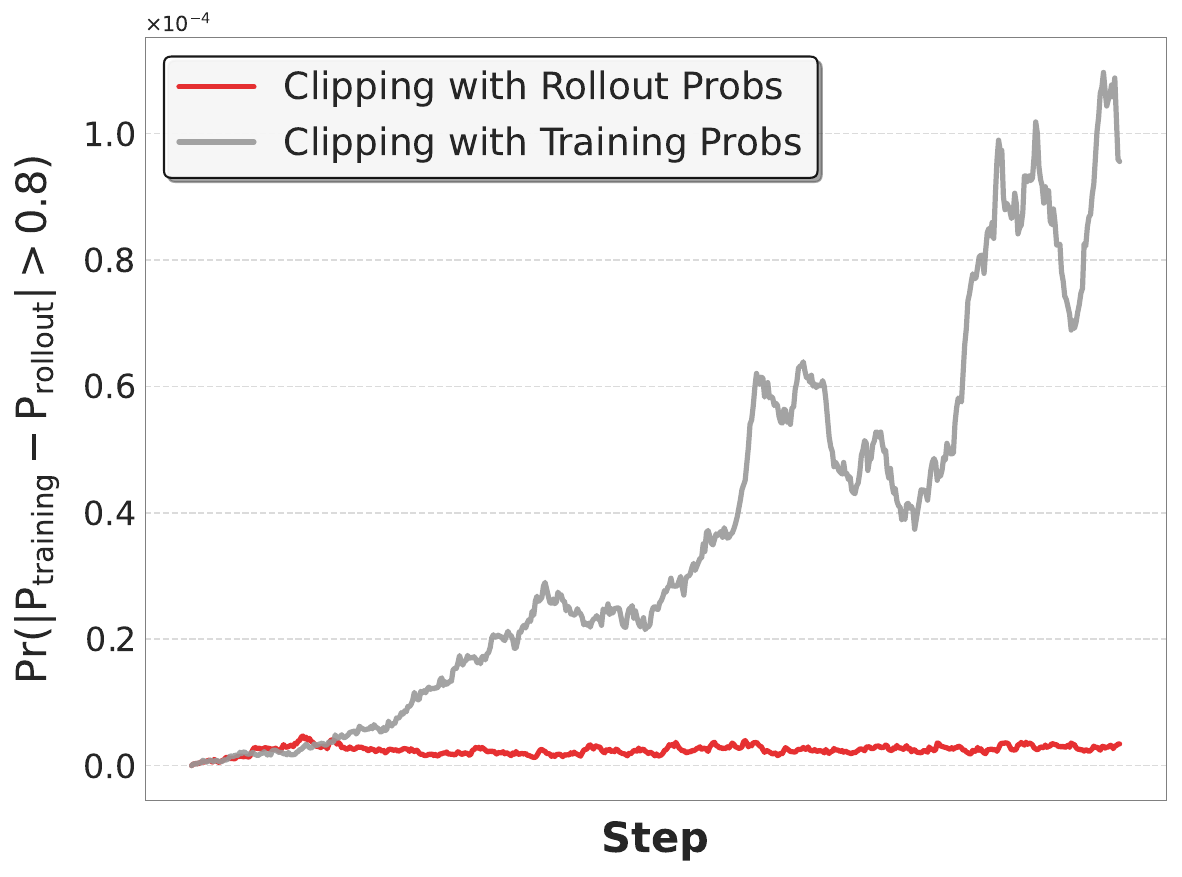}
        \label{fig:rl_alg_compare_diff}
    \end{subfigure}
    \caption{Comparison of PPO clip using rollout probabilities versus training probabilities. (Left) Reward. (Right) Proportion of tokens with an absolute training-inference probability difference greater than $0.8$.}
    \label{fig:rl_alg_compare}
\end{figure}

After systematic training-inference alignment, using rollout probabilities instead of recomputed training probabilities yields higher rewards in the later stages of training and maintains the training-inference disparity within a more stable range (as shown in Figure \ref{fig:rl_alg_compare}). In other words, under the premise of training-inference alignment, directly using rollout probabilities not only saves the time required for recomputing training probabilities but also further enhances the efficiency and stability of RL training.

\subsubsection{Experimental Results of Reinforcement Learning}
\begin{figure}[htb]
    \centering
    \begin{subfigure}[b]{0.32\textwidth}
        \centering
        \includegraphics[width=\textwidth]{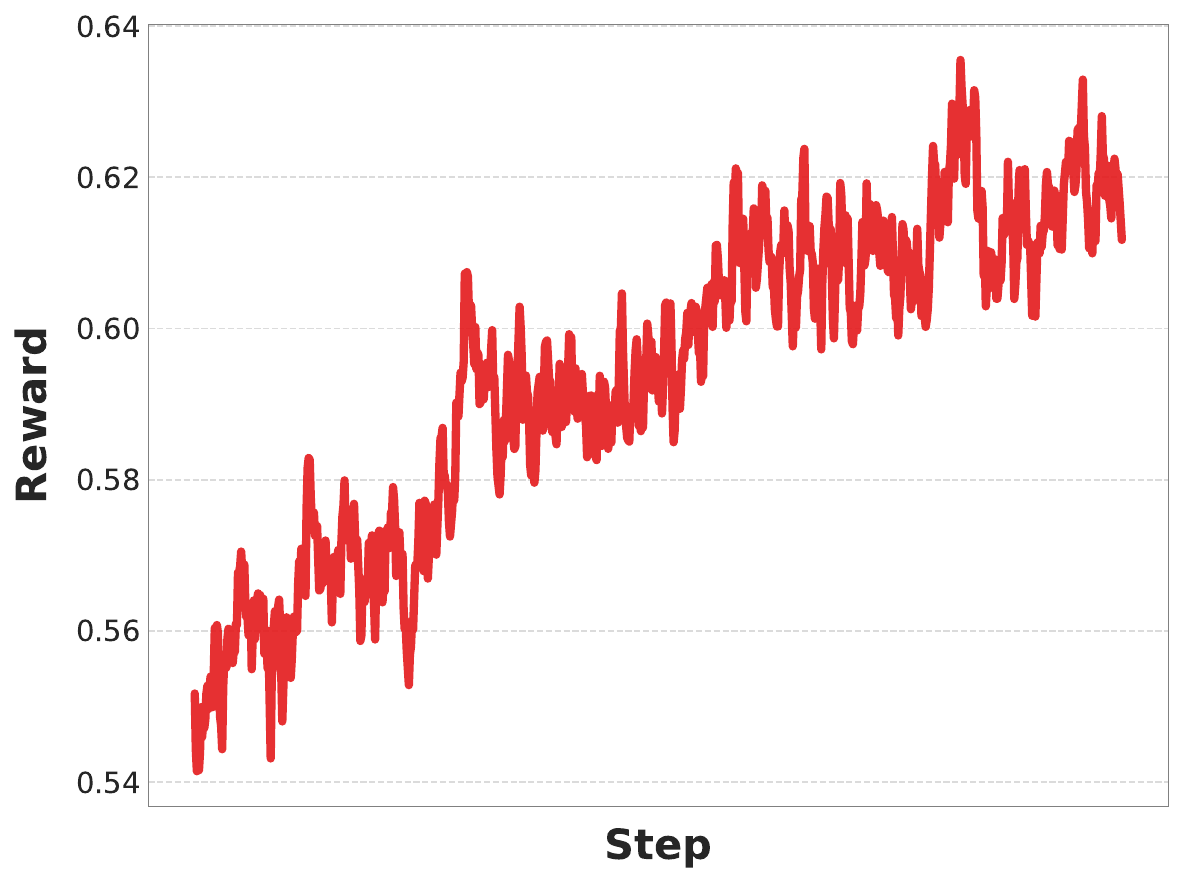}
        \caption{Training Reward.}
    \end{subfigure}
    \hfill
    \begin{subfigure}[b]{0.32\textwidth}
        \centering
        \includegraphics[width=\textwidth]{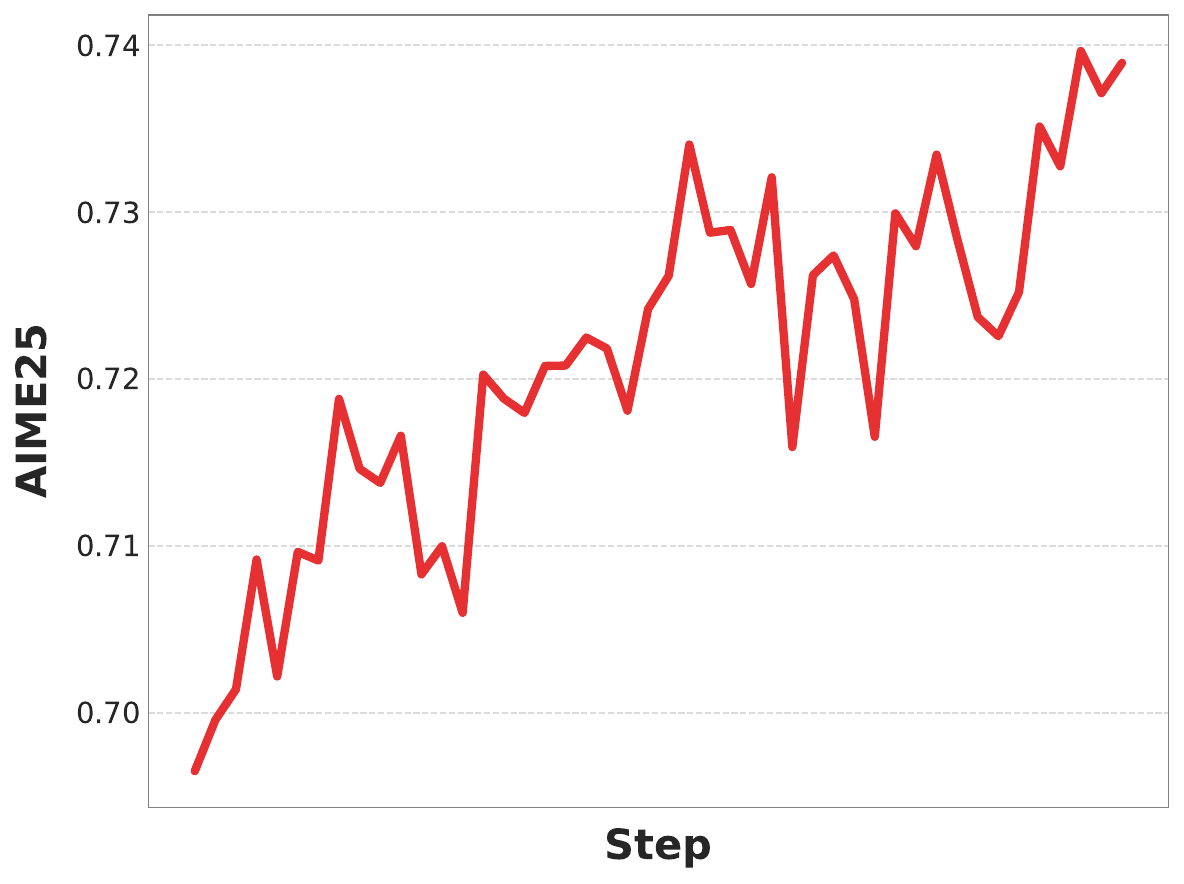}
        \caption{Test Score on AIME'25.}
    \end{subfigure}
    \hfill
    \begin{subfigure}[b]{0.32\textwidth}
        \centering
        \includegraphics[width=\textwidth]{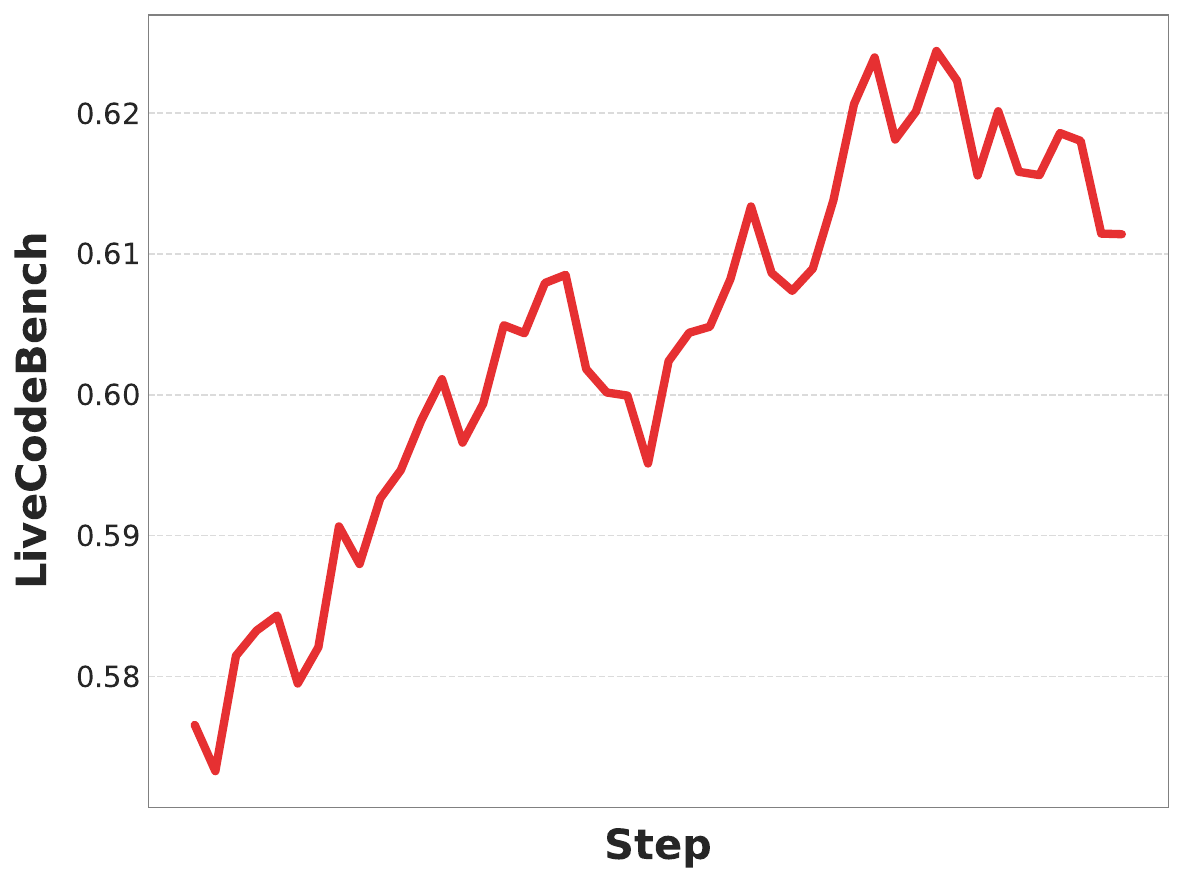}
        \caption{Test Score on LCB (sampled).}
    \end{subfigure}
    \caption{RL training curves of Ring-mini-linear-2.0.}
    \label{fig:rl_mini}
\end{figure}

Based on the aforementioned improvements, the model demonstrates consistent growth in both training reward and test score throughout the RL process, as shown in Figure \ref{fig:rl_mini}.

%% file: sections/6-evaluation.tex
\begin{table}[tbhp]
    \centering
    \caption{Performance of Ring-flash-mini-2.0 versus counterparts on various reasoning benchmarks.}
    \footnotesize
    \setlength{\tabcolsep}{1.6pt}
    \begin{tabular}{@{}c l | c  c  c   c@{}}
    \toprule
    & \multirow{2}{*}{\centering \textbf{Benchmark}}  & \textbf{Ring-Mini-}  & \textbf{Ring-Mini-}  & \textbf{Qwen3-8B-} & \textbf{GPT-OSS-}  \\
    & & \textbf{Linear-2.0}  & \textbf{2.0} & \textbf{Thinking} & \textbf{20B-Medium}\\
    \midrule
    \multirow{6}{*}{{\centering Mathematical Reasoning}}
     & AIME'24 & 79.95 & 79.69 & 79.27 & 77.86 \\
     & AIME'25 & 73.65 & 74.06 & 71.25 & 73.85 \\
     & OlympiadBench & 82.91 & 82.91 & 82.27 & 80.59 \\
     & CNMO'24 & 77.60 & 76.91 & 75.09 & 75.87 \\
     & LiveMathBench & 83.64 & 83.98 & 82.92 & 83.40 \\
     & TheoremQA & 69.69 & 70.09 & 68.81 & 66.81 \\
    \midrule
    \multirow{6}{*}{{\centering Agent \& Coding}}
     & Humaneval+ & 91.16 & 91.84 & 84.76 & 87.27 \\
     & MBPP+ & 79.70 & 80.79 & 79.63 & 79.86 \\
     & LiveCodeBench & 59.53 & 62.56 & 56.94 & 54.90 \\
     & CodeForces(Elo) & 83.84 & 84.80 & 73.31 & 82.25 \\
     & Spider & 79.18 & 77.64 & 79.41 & 78.81 \\
     & BFCL\_Live & 73.99 & 74.26 & 75.99 & 54.64 \\
    \midrule
    \multirow{5}{*}{{\centering General Reasoning}}
     & GPQA-Diamond & 65.69 & 68.24 & 62.00 & 65.53 \\
     & SciBench & 5.46 & 4.74 & 4.39 & 4.43 \\
     & DROP & 83.20 & 88.55 & 87.13 & 76.49 \\
     & MuSR & 77.21 & 75.99 & 76.92 & 76.53 \\
     & Multi\_LogiEval & 73.49 & 73.73 & 77.08 & 60.80 \\
    \bottomrule
    \end{tabular}
\label{tab:ring-mini-performance}
\end{table}

\section{Evaluation}\label{sec:evaluation}

We conduct a comprehensive evaluation of our Ring-linear series models, focusing on assessing the reasoning capabilities across multiple dimensions.

\subsection{Evaluation Setups}
This assessment is performed on a suite of 17 benchmarks across multiple reasoning dimensions:

\begin{itemize}

\item \textbf{Mathematical Reasoning}: AIME'24~\citep{MAA2024}, AIME'25~\citep{MAA2025}, OlympiadBench~\citep{he2024olympiadbench}, CNMO'24~\citep{CMS2024}, LiveMathBench~\citep{liu2024your}, TheoremQA~\citep{chen2023theoremqa}.

\item \textbf{Agent and Coding}: Humaneval+~\citep{evalperf}, MBPP+~\citep{evalperf}, LiveCodeBench~\citep{jain2024livecodebench}, CodeForces~\citep{Codeforces2024}, Spider~\citep{yu2018spider}, BFCL-Live~\citep{yan2024berkeley}.

\item \textbf{General Reasoning}: GPQA-Diamond~\citep{rein2024gpqa}, SciBench~\citep{wang2023scibench}, DROP~\citep{dua2019drop}, MuSR~\citep{sprague2023musr}, Multi-LogiEval~\citep{patel2024multi}.
\end{itemize}

We benchmark our Ring-linear series models against state-of-the-art counterparts of a similar parameter scale. 
Specifically, Ring-mini-linear-2.0 is compared against Ring-mini-2.0~\footnote{https://huggingface.co/inclusionAI/Ring-mini-2.0}, 
Qwen3-8B-thinking~\citep{yang2025qwen3}, and GPT-OSS-20B-Medium~\citep{agarwal2025gpt}. 
For Ring-flash-linear-2.0, the comparison includes Ring-Flash-2.0~\footnote{https://huggingface.co/inclusionAI/Ring-flash-2.0}, 
Qwen3-32B-Thinking~\citep{yang2025qwen3}, Gemini-2.5-Flash~\citep{comanici2025gemini}, GPT-OSS-120B-Medium~\citep{agarwal2025gpt}, Seed-OSS-36B-Instruct~\footnote{https://huggingface.co/ByteDance-Seed/Seed-OSS-36B-Instruct}, and Qwen3-Next-80BA3B-Thinking~\footnote{https://huggingface.co/Qwen/Qwen3-Next-80B-A3B-Thinking}.

\begin{table}[tbhp]
    \centering
    \caption{Performance of Ring-flash-linear-2.0 versus counterparts on various reasoning benchmarks.}
    \footnotesize
    \setlength{\tabcolsep}{1.3pt}
    \begin{tabular}{@{}c l | c  c  c c  c  c  c@{}}
    \toprule
    & \multirow{2}{*}{\centering \textbf{Benchmark}}  & \textbf{Ring-Flash-}  & \textbf{Ring-Flash-}  & \textbf{Qwen3-32B-} & \textbf{Gemini-2.5-} & \textbf{GPT-OSS-} & \textbf{Seed-OSS-} &\textbf{Qwen3-Next-} \\
    & & \textbf{Linear-2.0}  & \textbf{2.0} & \textbf{Thinking} & \textbf{Flash} & \textbf{120B-Medium} &  \textbf{36B-Instruct} &  \textbf{80BA3B-Thinking} \\
    \midrule
    \multirow{6}{*}{\parbox{1.8cm}{\centering Mathematical Reasoning}}
     & AIME'24 & 90.73 & 91.04 & 82.66 & 80.36 & 80.40 & 91.70 & 92.08 \\
     & AIME'25 & 86.51 & 86.98 & 75.47 & 72.00 & 80.00 & 84.70 & 87.80 \\
     & OlympiadBench & 87.36 & 88.10 & 84.69 & 83.41 & 82.32 & 87.06 & 87.80 \\
     & CNMO'24 & 84.98 & 85.07 & 78.21 & 82.38 & 79.95 & 91.75 & 85.33 \\
     & LiveMathBench & 88.11 & 87.84 & 86.85 & 85.86 & 85.89 & 88.70 & 88.08 \\
     & TheoremQA & 74.16 & 74.91 & 73.88 & 72.25 & 72.16 & 77.16 & 75.97 \\
    \midrule
    \multirow{6}{*}{\parbox{1.8cm}{\centering Agent \& Coding}}
    & Humaneval+ & 91.84 & 92.00 & 90.24 & 91.77 & 84.83 & 91.23 & 91.77 \\
    & MBPP+ & 81.02 & 81.28 & 82.28 & 80.42 & 79.99 & 80.89 & 81.55 \\
    & LiveCodeBench & 70.37 & 70.76 & 62.33 & 61.40 & 66.46 & 69.01 & 71.97 \\
    & CodeForces(Elo) & 90.24 & 90.23 & 84.25 & 81.59 & 89.67 & 83.78 & 89.77 \\
    & Spider & 80.86 & 81.70 & 81.00 & 77.27 & 78.71 & 78.16 & 82.95 \\
    & BFCL-Live & 75.51 & 75.22 & 76.34 & 75.26 & 60.22 & 61.05 & 77.17 \\
    \midrule
    \multirow{5}{*}{\parbox{1.8cm}{\centering General Reasoning}}
    & GPQA-Diamond & 74.49 & 75.25 & 68.40 & 82.80 & 73.10 & 71.40 & 77.20 \\
    & SciBench & 5.13 & 5.18 & 4.34 & 4.11 & 4.93 & 4.11 & 4.68 \\
    & DROP & 89.66 & 83.88 & 86.52 & 84.16 & 72.90 & 91.17 & 92.05 \\
    & MuSR & 84.05 & 85.11 & 78.21 & 84.98 & 82.57 & 82.06 & 81.03 \\
    & Multi\_LogiEval & 75.21 & 76.18 & 77.01 & 77.97 & 73.96 & 77.90 & 72.62\\
    \bottomrule
    \end{tabular}
\label{tab:ring-flash-performance}
\end{table}

\subsection{Main Results}

The performance results of Ring-mini-linear-2.0 and its comparison methods are summarized in Table~\ref{tab:ring-mini-performance}. It is evident from the table that the Ring-mini-linear-2.0 model, despite its compact scale of only 1.6B activated parameters, demonstrates performance that is comparable to its counterpart models on various reasoning tasks.

Table~\ref{tab:ring-flash-performance} summarizes the benchmark results of our Ring-flash-linear-2.0 model alongside several key counterparts. 
The results indicate that despite its compact design, Ring-flash-linear-2.0 delivers performance that is highly comparable to these compared models across a diverse range of reasoning dimensions, demonstrating the model's comprehensive and robust reasoning capabilities.

%% file: sections/7-conclusion.tex
\section{Conclusion}
This technical report presents models based on the Ring-linear hybrid architecture, specifically Ring-mini-linear-2.0 and Ring-flash-linear-2.0. By integrating a linear attention mechanism with high-performance FP8 fused kernels--LingHe, we have significantly enhanced both the training and inference efficiency of the models. Moreover, a well-aligned RL system has contributed to more stable training, further elevating the model's performance ceiling.
However, we also recognize certain limitations in the current models. 
For instance, to maintain model effectiveness, the linear attention module maintains an identical attention head count for Q, K, and V, which brings heavy memory overhead. 
Furthermore, the remaining softmax attention modules introduce additional computational bottlenecks, impacting the overall efficiency of the hybrid architecture.
Moving forward, we will continue to explore more efficient model architectures to better balance performance and efficiency.